\documentclass[lettersize,journal]{IEEEtran}
\IEEEoverridecommandlockouts

\pdfoutput=1

\usepackage{graphicx}
\usepackage{epsfig} 
\usepackage{amsmath} 

\usepackage{amssymb}  
\usepackage{amsthm}
\DeclareMathAlphabet{\altmathcal}{OMS}{cmsy}{m}{n}
\usepackage{mathrsfs}
\usepackage{bm}
\usepackage{float}
\usepackage{bbm}
\usepackage{color}
\usepackage{lipsum}
\usepackage{placeins}
\usepackage[ruled,vlined,linesnumbered]{algorithm2e}
\usepackage[colorlinks=true,linkcolor=black,anchorcolor=black,citecolor=black,filecolor=black,menucolor=black,runcolor=black,urlcolor=cyan]{hyperref}
\usepackage{wrapfig}
\usepackage[table,xcdraw,dvipsnames]{xcolor}
\usepackage{multirow}
\usepackage{mathtools}
\usepackage{algpseudocode}
\usepackage{xspace}
\usepackage{etoolbox}
\usepackage{tikz}
\usetikzlibrary{calc}

\hypersetup{
    colorlinks=true,
    linkcolor=blue,
    filecolor=magenta,      
    urlcolor=cyan,
    pdftitle={SPLANNING},
    pdfpagemode=FullScreen}

\usepackage{outlines}
\let\labelindent\relax
\usepackage{enumitem}
\setenumerate[1]{label=\Roman*.}
\setenumerate[2]{label=\Alph*.}
\setenumerate[3]{label=\arabic*.}
\setenumerate[4]{label=\roman*.}

\usepackage{marginnote}
\usepackage[normalem]{ulem}
\usepackage{cancel}          

\usepackage[footnotesize]{caption}
\usepackage{subcaption}

\usepackage{pifont}

\usepackage{booktabs}

\makeatletter
\patchcmd{\@makecaption}
  {\scshape}
  {}
  {}
  {}
\makeatother
\usepackage{changes}

\newcommand{\green}[1]{{\color{ForestGreen} #1}}
\definecolor{StartBlue}{HTML}{2E8ED3}
\definecolor{EndGreen}{HTML}{2CBB00}
\definecolor{ReachPurple}{HTML}{7420CC}

\newcommand{\stuck}[1]{{\color{orange} #1}}
\newcommand{\crash}[1]{{\color{red} #1}}
\newcommand{\success}[1]{#1}

\newif\ifreviewson
\newif\ifcommentson
\commentsonfalse
\reviewsonfalse

\newcommand{\blueReview}[1]{\ifreviewson{\color{ForestGreen} #1} \else #1 \fi}

\newcounter{FixCount}
\addtocounter{FixCount}{1}

\newcommand{\soutsmart}[1]{
  \ifmmode
    \cancel{#1}%
  \else
    \sout{#1}%
  \fi}
  
\providecommand{\Review}[3]{%
  \ifreviewson
    {\green{{#2}}}%
  \else
    #2%
  \fi}
  
\providecommand{\methodname}{\text{SPLANNING}\xspace}
\providecommand{\sparrows}{\text{SPARROWS}\xspace}
\providecommand{\armtd}{\text{ARMTD}\xspace}

\providecommand{\mpot}{\text{MPOT}\xspace}
\providecommand{\chomp}{\text{CHOMP}\xspace}

\providecommand{\trajopt}{\text{TrajOpt}\xspace}
\providecommand{\curobo}{\text{cuRobo}\xspace}
\providecommand{\catnips}{\text{CATNIPS}\xspace}
\providecommand{\splatnav}{\text{Splat-Nav}\xspace}
\providecommand{\catnipsstar}{\text{CATNIPS$^\ast$}\xspace}
\providecommand{\splatnavstar}{\text{Splat-Nav$^\ast$}\xspace}
\providecommand{\nerfnav}{\text{NeRF-Nav}\xspace}

\newcommand{\norm}[1]{\left\Vert#1\right\Vert}
\newcommand{\abs}[1]{\left\vert#1\right\vert}

\newtheorem{defn}{Definition}
\newtheorem{rem}[defn]{Remark}
\newtheorem{lem}[defn]{Lemma}

\newtheorem{assum}[defn]{Assumption}

\newtheorem{thm}[defn]{Theorem}

\newcommand{\regtext}[1]{\mathrm{\textnormal{#1}}}

\newcommand{\ts}[1]{\textsuperscript{#1}}

\renewcommand{\det}[1]{\lvert#1\rvert}

\providecommand{\R}{\ensuremath \mathbb{R}}
\providecommand{\N}{\ensuremath \mathbb{N}}

\providecommand{\diag}{\texttt{diag}}

\providecommand{\tplan}{t_p}

\providecommand{\opt}{\texttt{(Opt)}}
\providecommand{\splanningopt}{\texttt{(Splanning-Opt)}}

\providecommand{\splanningoptref}{\hyperref[eq:splanning_opt_cost]{\splanningopt{}}}

\providecommand{\world}{W}
\providecommand{\workspace}{W_s}

\newcommand{\nq}{n_q}
\newcommand{\ns}{n_s}
\newcommand{\nt}{n_t}

\newcommand{\Nq}{ N_q }
\newcommand{\Ns}{ N_s }

\providecommand{\T}{\mathcal{T}}

\providecommand{\E}{\mathbb{E}}

\newcommand{\numop}[1]{{\mathrm{\textnormal{\texttt{#1}}}}}

\newcommand{\qlim}{q_{j,\regtext{lim}}}
\newcommand{\dqlim}{\dot{q}_{j,\regtext{lim}}}

\newcommand{\timestep}{\Delta t}

\newcommand{\conv}{co}

\providecommand{\x}{x}
\renewcommand{\L}{\mathcal{L}}

\providecommand{\tfin}{t_\text{f}}
\providecommand{\T}{\ensuremath T}

\newcommand{\zeros}{\textit{0}}

\newcommand{\pow}[1]{\altmathcal{P}\!\left(#1\right)}

\providecommand{\sfo}{\altmathcal{SFO}}

\newcommand{\SFO}{{Spherical Forward Occupancy}\xspace}

\newcommand{\pz}[1]{\mathbf{#1}}

\newcommand{\pzi}[1]{\pz{ #1 }(\pz{T_i};\pz{K})}

\newcommand{\pzki}[1]{\pz{ #1 }(\pz{T_i};k)}

\newcommand{\pzqki}{\pzki{q}}

\newcommand{\pzqji}{\pzi{q_j}}
\newcommand{\pzqdji}{\pzi{\dot{q}_j}}

\newcommand{\pzqjki}{\pzki{q_j}}

\newcommand{\tvari}{x_{t_{i}}}

\newcommand{\q}{q(t)}

\newcommand{\qj}{q_j(t)}
\newcommand{\ql}{q_l(t)}
\newcommand{\qdj}{\dot{q}_{j}(t)}

\newcommand{\qkj}{q_j(t; k)}
\newcommand{\qdkj}{\dot{q}_{j}(t; k)}

\newcommand{\FK}{\regtext{\small{FK}}}

\newcommand{\Btwo}{B_{2}}

\providecommand{\FO}{\regtext{\small{FO}}}

\newcommand{\FOjik}{FO_{j}(\pzqki)}

\newcommand{\Sjq}{S_j(\q)}

\newcommand{\Sjpq}{S_{j+1}(\q)}

\newcommand{\Sbarjimk}{{\bar{S}_{j,i,m}(\pzqki)}}

\newcommand{\TCjik}{TC_{j}(\pzqki)}

\newcommand{\transmit}{\altmathcal{T}}
\newcommand{\ray}{\varphi}

\newcommand{\radiance}{\altmathcal{L}}
\newcommand{\raydir}{{v_\ray}}
\newcommand{\raydirvar}{\mathbf{V}}
\newcommand{\rayproc}{\mathbf{\Phi}}
\newcommand{\rayorig}{{o_\ray}}

\renewcommand{\S}{\mathbb{S}}

\newcommand{\pr}{\mathrm{P}}
\newcommand{\erf}{\mathrm{erf}}
\newcommand{\bound}{\mathcal{H}}

\newcommand{\best}[1]{\textbf{#1}}
\newcommand{\ru}[1]{\underline{#1}}

\usepackage[
    style=ieee,
    doi=false,
    isbn=false,
    url=false,
    eprint=false,
    backend=biber,
    natbib=true,
    citestyle=numeric-comp
    ]{biblatex}
    
\bibliography{references}

\def\anonymous{1}

\begin{document}

\title{Let's Make a Splan: Risk-Aware Trajectory Optimization in a Normalized Gaussian Splat}

\ifthenelse{\equal{\anonymous}{2}}{
\author{Author Names Omitted for Anonymous Review. Paper-ID [add your ID here]}
}{
\author{Jonathan Michaux*, Seth Isaacson*, Challen Enninful Adu, Adam Li, Rahul Kashyap Swayampakula,\\
Parker Ewen, Sean Rice, Katherine A. Skinner, and Ram Vasudevan
\thanks{*Denotes equal contribution.}
\thanks{Jonathan Michaux, Seth Isaacson, Challen Enninful Adu, Adam Li, Rahul Kashyap Swayampakula, Parker Ewen, Sean Rice, Katherine A. Skinner and Ram Vasudevan are with the Department of Robotics, University of Michigan, Ann Arbor, MI 48109. \texttt{\{jmichaux, sethgi, enninful, adamli, rahulswa, pewen, seanrice, kskin, ramv\}@umich.edu}.}
\thanks{This work was funded by MCity, University of Michigan.}
}}

\maketitle

\begin{abstract}
Neural Radiance Fields and Gaussian Splatting have recently transformed computer vision by enabling photo-realistic representations of complex scenes. However, they have seen limited application in real-world robotics tasks such as trajectory optimization. This is due to the difficulty in reasoning about collisions in radiance models and the computational complexity associated with operating in dense models. This paper addresses these challenges by proposing SPLANNING, a risk-aware trajectory optimizer operating in a Gaussian Splatting model. This paper first derives a method to rigorously upper-bound the probability of collision between a robot and a radiance field. Then, this paper introduces a normalized reformulation of Gaussian Splatting that enables efficient computation of this collision bound. Finally, this paper presents a method to optimize trajectories that avoid collisions in a Gaussian Splat. Experiments show that SPLANNING outperforms state-of-the-art methods in generating collision-free trajectories in cluttered environments. The proposed system is also tested on a real-world robot manipulator.
A project page is available at \href{https://roahmlab.github.io/splanning}{https://roahmlab.github.io/splanning}.
\end{abstract}
\begin{IEEEkeywords}
Collision Avoidance, Motion and Path Planning, 3D Gaussian Splatting.
\end{IEEEkeywords}

\section{Introduction}\label{sec:intro}
For a robot to safely navigate its environment, it must understand the scene geometry it operates within.
This understanding must include a detailed model of the scene and a method to reason about collisions with the environment.
Radiance field representations, such as Neural Radiance Fields (NeRFs \cite{mildenhall2020nerf}) and Gaussian Splatting \cite{kerbl20233d}, have recently emerged as powerful methods for building detailed models of the scene.
A radiance field is a five-dimensional function that maps a 3D point and viewing direction to an RGB color and volume rendering opacity.
This function is then integrated along camera rays to approximate the image formation process.
NeRFs use neural networks to learn the parameters of a radiance field, while Gaussian Splatting models use a set of 3D Gaussian functions.
Over the past several years, radiance field representations have marked a paradigm shift in computer vision, with wide-ranging impacts on scene reconstruction \cite{zhang2022nerfusion, pumarola2021d}, novel view synthesis \cite{mildenhall2020nerf}, 3D tracking \cite{yuan2021star, xu2021h}, and more.

\begin{figure}[t]
  \centering
  \includegraphics[width=0.9\columnwidth]{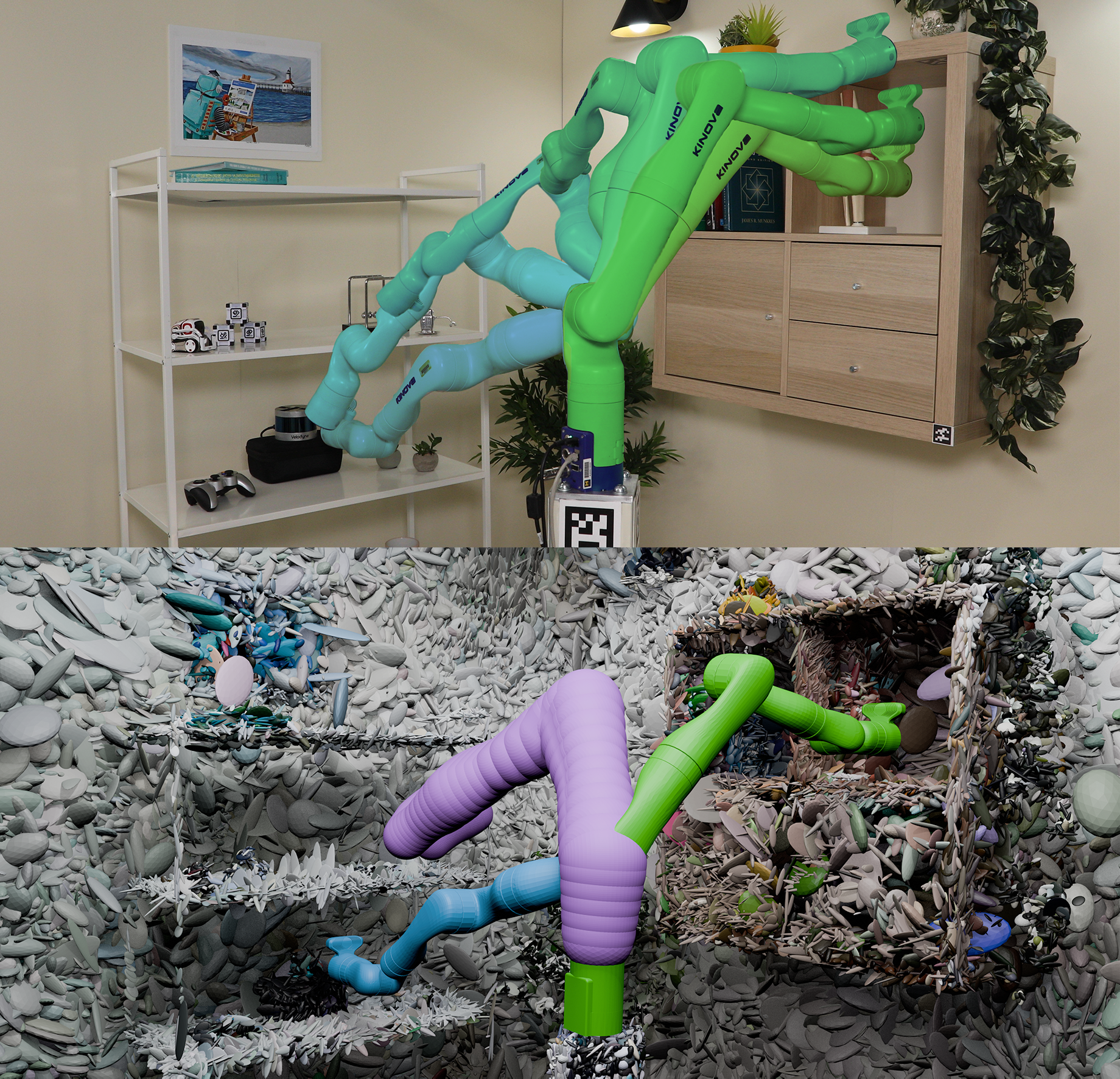}
  \caption{\methodname constructs risk-aware trajectories in a Gaussian Splatting map in real-time. 
  The top image shows a real-world scene that a 7DOF serial manipulator must plan through, starting at the \textcolor{StartBlue}{blue} configuration (left) and ending at the \textcolor{EndGreen}{green} configuration (right). 
  The scene is represented as a normalized 3D Gaussian Splat. 
  Then, in real-time, \methodname solves an optimization problem that constrains the probability that the robot's forward occupancy
  (bottom center, \textcolor{ReachPurple}{purple}) collides with the scene.}
  \label{fig:fig1}
  \vspace{-0.5cm}
\end{figure}

The robotics research community has begun trying to integrate these models for robotic tasks such as localization \cite{li2023nerfpose}, mapping \cite{sucar2021imap, isaacson2023loner}, and navigation \cite{adamkiewicz2021nerfnav}. 
\Review{A key strength of radiance models for robotics applications is that they continuously represent the environment.
This differs from conventional discrete representations in robotics, such as point clouds and occupancy grids.
However, reasoning rigorously about collisions in these continuous environments is challenging.
While existing planners offer practical solutions to this problem, such as discretizing the robot body \cite{adamkiewicz2021nerfnav}, discretizing the map before planning \cite{chen2023catnips}, or treating the confidence ellipsoids of the Gaussians in a Gaussian Splat as obstacles \cite{chen2024splatnav}, work remains to fully exploit the continuous nature of radiance field models.}
{A key strength of radiance models, such as 3D Gaussian Splats (3DGS), is that they represent the 3D scene using a continuous basis set. 
This differs from conventional discrete representations in robotics, such as point clouds and occupancy grids.
In particular, point clouds do not inherently encode surface or object connectivity because they are a sampled representation of the environment.
    As a result, it's unclear how to derive a probabilistic interpretation for the occupancy of a continuous portion of the scene from a point cloud.
    Though voxel grids do have a direct volumetric and probabilistic interpretation \cite{hornung13auro},
    the resolution of the map must be selected a-priori, which presents a strong tradeoff between fidelity and computational demands.
    In addition, to ensure safe planning, voxel-based methods typically employ obstacle buffering to account for the robot's geometry.
    This is straightforward to perform if the robot's footprint can be modeled as a point or sphere; however, this buffering operation is challenging to perform for articulated robots without unnecessarily restricting free space \cite{lozano1987simple}.
    On the other hand, just as with point clouds, one could process the voxel grid to create a surface model such as a signed distance field.
    However, this introduces additional computational overhead and potential inaccuracies in collision checking \cite{eriksson2024fast}.
    Despite representing the scene using a continuous basis set, reasoning rigorously about collisions in 3DGS representations is challenging. 
    While existing planners offer practical solutions to this problem, such as discretizing the robot body \cite{adamkiewicz2021nerfnav}, discretizing the map before planning \cite{chen2023catnips}, or treating the confidence ellipsoids of the Gaussians in a Gaussian Splat as obstacles \cite{chen2024splatnav}, work remains to fully exploit the continuous nature of radiance field models.
    }{1-I}

This paper extends the literature on motion planning in radiance fields by proposing a real-time, receding-horizon trajectory optimization algorithm called \methodname.
This paper's key contributions are: 
1) A rigorous definition and derivation of rigid body collision within a radiance field model, starting directly from the rendering equation;
\reversemarginpar
\Review{ 2) a computationally efficient approach to upper-bound the probability of collision within a Gaussian Splatting model;}{2) a computationally efficient approach to upper-bound the probability of collision within a Gaussian Splatting model that can be incorporated into a real-time risk-aware trajectory planner;}{2-I}
\normalmarginpar
3) a re-formulation of Gaussian Splatting that normalizes the 3D Gaussians to ensure the correctness of the collision probabilities
\Review{; 4) a novel risk-aware trajectory planner for robot manipulators}{}{}. 
Simulation and hardware experiments illustrate that the risk-aware planner solves challenging tasks in real-time.

\textbf{Relationship to Prior Work:}
\methodname builds upon prior work entitled Safe Planning for Articulated Robots Using Reachability-based Obstacle Avoidance With Spheres (\sparrows) \cite{michaux2024sparrows}.
The prior work develops a trajectory optimization algorithm that leverages a novel sphere-based reachable set that overapproximates the swept volume of a serial robot manipulator.
At runtime, \sparrows uses this representation to enforce collision-avoidance constraints with obstacles of known geometry.
In contrast, the present work introduces a novel chance constraint to facilitate planning in scenes with arbitrary geometry modeled by radiance fields.

\section{Related Work}\label{sec:related_work}
\methodname is an algorithm combining trajectory optimization, reachability analysis for robot safety, and radiance fields for 3D scene representation.
We discuss the relevant literature here.

\subsection{Trajectory Optimization}
To generate safe motion plans, state-of-the-art trajectory optimizers such as \chomp\cite{Zucker2013chomp}, \trajopt \cite{Schulman2014trajopt}, \mpot \cite{le2023mpot}, and \curobo \cite{sundaralingam2023curobo} model the robot or the environment with simple geometric primitives such as spheres \cite{duenser2018manipulation, gaertner2021collisionfree}, ellipsoids \cite{brito2020model}, capsules \cite{dube2013humanoids,khoury2013humanoids}, or convex polygons and perform collision-checking along a given trajectory at discrete time instances.
\chomp represents the robot as a collection of discrete spheres and avoids collisions using a signed distance field to maintain a safety margin with the environment.
\trajopt uses the support mapping of convex shapes to represent the environment obstacles and the robot.
Then, the signed distance (positive distance and penetration depth) between two convex shapes is computed by the Gilbert-Johnson-Keerthi \cite{gtk1988} and Expanding Polytope Algorithms \cite{bergen2001penetration}.
\mpot represents the robot geometry and environment obstacles as a collection of spheres and implements a collision-avoidance cost using an occupancy map \cite{le2023mpot}.
Using a gradient-free approach, \mpot optimizes a batch of smooth trajectories and selects the one with the lowest cost.
Similarly, a recent method called \curobo \cite{sundaralingam2023curobo} represents the robot as a collection of spheres and solves multiple trajectory optimization problems in parallel on the GPU to identify a collision-free path.
\Review{The major drawback of these approaches is that the resulting trajectories are not guaranteed to be safe, as collision avoidance is enforced only as a soft penalty in the cost function.}
{In each of these approaches, collision avoidance is enforced using a soft penalty in the cost function. 
The major drawback of these approaches is that the resulting trajectories are not guaranteed to be safe.}{4-II}
\Review{Furthermore, these approaches require explicit representations of scene geometry, such as zonotopes, point clouds, or 3D occupancy grids, which can be challenging to construct in real-world robotics applications.
In contrast, \methodname enforces collision avoidance as a constraint and operates directly in a learned Gaussian Splatting model.}
{Furthermore, these approaches require explicit representations of scene geometry, such as zonotopes, point clouds, or 3D occupancy grids.
Each of these methods suffers from a fundamental limitation: zonotopes are challenging to construct from sensor data, point clouds lack a clear volumetric and probabilistic interpretation, and 3D occupancy grids' computational requirements grow cubically with the resolution and size of the map.
In contrast, \methodname enforces collision avoidance as a constraint and operates directly in a learned Gaussian Splatting model.
In addition to being constructible directly from sensor data, Gaussian Splats allow for photorealistic rendering of the scene.}{3-II}

\begin{figure*}[t!h!]
    \centering
    \includegraphics[width=1.0\textwidth]{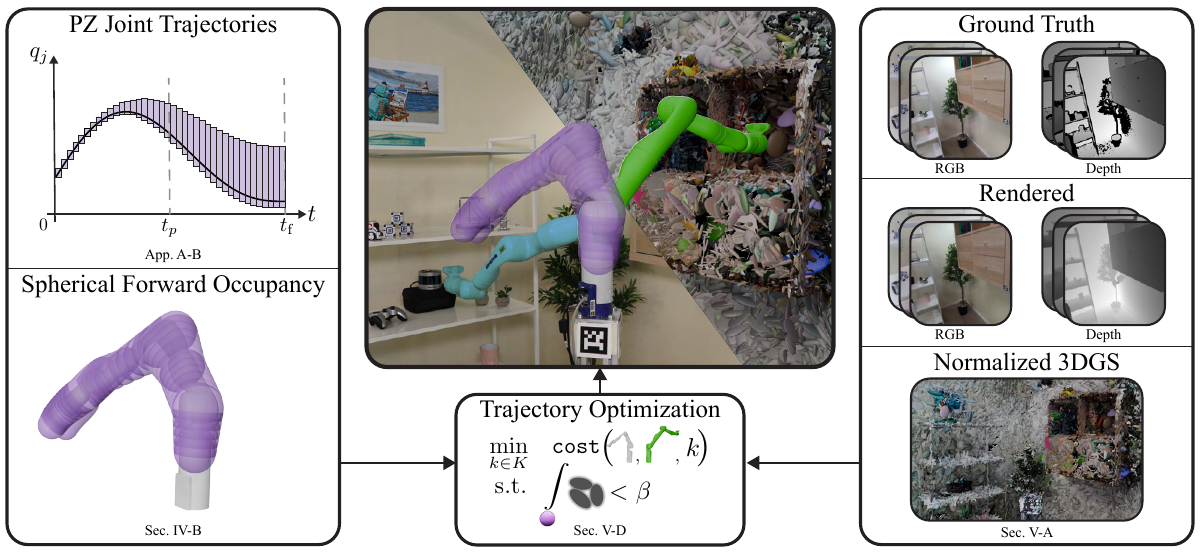}
    \caption{\methodname optimizes trajectories in a Normalized 3D Gaussian Splat given a start configuration \textcolor{StartBlue}{(blue)} and goal configuration \textcolor{EndGreen}{(green)}.  
    Offline, a Normalized 3D Gaussian Splat is constructed to represent the scene geometry \Review{}{(Sec. \ref{subsec:approach_splatting}, bottom right).}{29-III-1}
    \Review{}{Online, a family of parameterized trajectories (App. \ref{app:modeling_trajectory}) is partitioned into a finite set of intervals (App. \ref{app:reachability}, top left).}{}
    \Review{Then, during online planning, a constraint bounds the probability that the Spherical Forward Occupancy \textcolor{ReachPurple}{(purple)} intersects with the scene, as represented by a Normalized 3D Gaussian Splat.}
    {Then, for each time interval, the Spherical Forward Occupancy ($\sfo$) \textcolor{ReachPurple}{(purple)} is computed as an overapproximation of the robot's swept volume (Sec. \ref{subsec:modeling_arm_occupancy}, bottom left).
    Finally, during online trajectory optimization (Sec. \ref{subsec:approach_motion_planning}, bottom middle), a novel constraint (Sec. \ref{subsec:approach_collision}--\ref{subsec:approach_integral_evaluation}) bounds the probability that the $\sfo$ intersects with the scene, as represented by a Normalized 3D Gaussian Splat}{}.
    }
    \label{fig:opening_figure}
    \vspace*{-0.3cm}
\end{figure*}

\subsection{Reachability Analysis}
Reachability-based Trajectory Design (RTD) \cite{kousik2017safe} is a recent approach to real-time motion planning that generates provably\Review{}{-}{}safe trajectories in a receding-horizon fashion.
At runtime, RTD constructs reachable sets that overapproximate all possible robot configurations corresponding to a pre-specified continuum of parameterized trajectories.  
RTD then solves a nonlinear optimization problem to select a feasible trajectory such that the robot's motion is guaranteed to be collision-free.
If a feasible trajectory is not found, RTD brings the robot safely to a stop by executing a fail-safe braking maneuver.
Unlike traditional trajectory optimization methods, RTD constructs reachable sets such that obstacle-avoidance constraints are satisfied in \emph{continuous-time}.
Recent extensions of RTD have demonstrated real-time, certifiably-safe motion planning for robotic arms \cite{holmes2020armtd, michaux2023armour, michaux2023rdf, brei2024waitr, michaux2024sparrows} and mobile robots \cite{liu2022refine, michaux2024redefined}.
Probabilistic extensions of RTD have also been proposed \cite{liu2023radius}, which provide safety guarantees in uncertain environments.
However, these reachability-based methods assume ground-truth knowledge of obstacle geometry or that probability density functions on the locations of obstacles are provided.
\methodname extends the literature on reachability analysis by presenting a reachability-based planner that operates in a radiance field model. 


\subsection{Radiance Fields}
Neural Radiance Fields (NeRFs) were first introduced to address the problem of novel view synthesis \cite{mildenhall2020nerf}.
NeRFs and their variants use neural networks to estimate the radiance emitted by a scene point when viewed from a given direction.
Since their introduction, NeRFs have found many use cases in robotics \cite{wang2024nerfRoboticsSurvey} including navigation \cite{adamkiewicz2021nerfnav, katragadda2024nerfvins}, pose estimation \cite{yen2021inerf, li2023nerfpose}, manipulation \cite{ichnowski2021dexnerf}, and SLAM \cite{sucar2021imap, isaacson2023loner}.
\nerfnav \cite{adamkiewicz2021nerfnav}, an early effort toward safe planning in radiance fields, approximates the robot as a set of points and then avoids collisions by integrating the NeRF density along the path traced by each point.
A later work, \catnips~\cite{chen2023catnips}, instead presents a framework for relating NeRFs to Poisson Point Processes.
\catnips then introduces a method to convert a NeRF to an occupancy grid to plan robot motions.

While NeRFs use a neural network to model image formation, Gaussian Splatting methods learn a similar representation using Gaussian basis functions \cite{zwicker2002ewa}.
Recently, Kerbl et al. introduced 3D Gaussian Splatting (3DGS) \cite{kerbl20233d}, which learns the parameters of un-normalized 3D Gaussians via gradient descent.
3DGS offers high-speed rendering and a training method compatible with modern Graphics Processing Units (GPUs).
However, similar to NeRFs, only limited efforts have been made towards using 3DGS for real-time motion planning.
\Review{A recent work called \splatnav \cite{chen2024splatnav} leverages the fact that 3DGS only renders Gaussians where the 99\% confidence interval overlaps with the camera's view frustum.
This allows \splatnav to ignore all other Gaussians while generating motion plans within safe corridors that avoid collisions with the 99\% confidence intervals.
Still, \splatnav uses sampling-based methods to check collisions.
In addition, \splatnav's collision checking is not differentiable.}{\splatnav is a real-time robot navigation pipeline designed specifically for Gaussian Splatting representations. 
It works by constructing safe polytope corridors through the environment by deriving the distance between a robot described by an ellipsoid while traveling along a straight line and another ellipsoid that corresponds to a Splat within the scene. 
It then optimizes Bezier curve trajectories within these corridors, ensuring they remain collision-free. 
This approach is validated on quadrotors.

Unfortunately, extending this approach to robots such as manipulators presents significant challenges. 
First, motion planning within polytopic constraints is well-understood for quadrotors due to their simpler dynamics and direct workspace constraints, but it becomes considerably more complex for articulated manipulators.
Their high-dimensional, nonlinear configuration spaces create intricate, often non-convex, mappings from workspace polytopes to configuration space constraints.
Second, \splatnav's distance computation approach is designed specifically for straight-line movements and would not be immediately applicable when performing trajectory optimization over arbitrary trajectories (e.g., non-straight lines in the workspace). 
Manipulator planning typically requires considering paths through configuration space rather than linear segments in workspace, necessitating a different mathematical formulation for collision avoidance.
Finally, the Splat-Nav constraint is implemented using a bisection search, complicating its integration into differentiable trajectory optimizers.}{6-II}

In contrast, this paper proposes a novel re-formulation of 3DGS enabling collision checking that is computationally simple and fully differentiable, allowing it to be incorporated into a trajectory optimizer and gradient-based learning methods.
Notably, we demonstrate that our collision avoidance representation is more accurate than existing methods.
Further, we describe a planning framework that works with serial robot manipulators, unlike prior risk-aware planners for radiance field models that operate on quadrotors \cite{adamkiewicz2022vision, chen2023catnips, chen2024splatnav}.


\section{Proposed Method Overview}\label{sec:overview}
As illustrated in Fig.~\ref{fig:opening_figure}, \methodname computes risk-aware trajectories using a visual scene representation.
The key insight behind \methodname is the combination of reachability analysis with a novel Normalized 3D Gaussian Splat.
This combination allows \methodname to constrain the probability of collision between the robot's reachable set and the scene.

\subsection{Notation}
We briefly outline the notation used throughout the rest of the document.

\subsubsection{Sets and Functions}
$n_A \in \N$ represents an arbitrary constant, while $N_A = \{1,\hdots,n_A\} \subset \N$ is a set.
Capital Roman letters $X$ denote sets, and $\conv(X)$ denote the convex hull of $X$.
$f(x)$ denotes a function of $x$, while $f[g]$ denotes a functional of the function $g$.

\subsubsection{Probabilities}
A probability space is denoted $(\Omega, \mathcal{F}, \mathcal{P})$ with sample space $\Omega$, event space $\mathcal{F}$, and probability measure $\mathcal{P}$. 
A random variable is denoted $\mathbf{X}: \Omega \to S$ for some measurable space $S$.
For a random event $E$, the probability that $E$ occurs is denoted $\pr(E)$.

\subsubsection{Robot Descriptions}
$q(t;k)$ and $\dot{q}(t,k)$ represent a robot's time-dependent configuration and velocity, respectively.
$k \in K \subset \R^{n_k}$ parameterizes a continuum of trajectories, and $t \in T \subset \R$ specifies the time horizon.
The forward occupancy of a robot while following a trajectory $q(t;k)$ is denoted $\FO(q(t;k))$.

\subsection{Overview of Optimization}

At runtime, \methodname computes probabilistically safe trajectories by solving an optimization problem that limits the probability of the robot's forward occupancy intersecting the environment.
\methodname tries to solve the following optimization problem in a receding-horizon manner:
\begin{align}
    \label{eq:optcost}
    &\underset{k\in K}{\min} &&\texttt{cost}(k) \\
    \label{eq:optpos}
    &&& q(t; k) \in [\qlim^-, \qlim^+]  &\forall t \in T \\
    \label{eq:optvel}
    &&& \dot{q}(t; k) \in [\dqlim^-, \dqlim^+]  &\forall t \in T\\
    \label{eq:optcolcon}
    &&& \pr\left(\FO(q(t; k)) \cap \mathscr{E} \neq \emptyset\right) < \beta   &\forall t \in T
\end{align}

The cost function \eqref{eq:optcost} is a user-defined objective, such as bringing the robot close to a desired goal.
Input constraints enforce limits on joint positions \eqref{eq:optpos} and velocities \eqref{eq:optvel}.
Finally, \eqref{eq:optcolcon} is a safety constraint ensuring that the probability of the robot's forward occupancy $\FO(q(t;k))$ intersecting the environment $\mathscr{E}$ stays below the risk threshold $\beta$.

The rest of this paper is organized as follows.
Section~\ref{sec:robot_representations} details the representation of the robot and outlines how the forward occupancy of the robot is overapproximated.
Section~\ref{sec:method} provides a detailed description of \methodname, including the Normalized 3D Gaussian Splatting used to represent the scene and a method for overapproximating the risk constraint in \eqref{eq:optcolcon}.
Section~\ref{sec:results} evaluates \methodname both in simulation and on hardware, and Section~\ref{sec:conclusion} concludes and discusses avenues for future work.

\section{Robot Representation}
\label{sec:robot_representations}

This section describes the construction of a sphere-based safety representation for a serial robotic manipulator.
Section~\ref{subsec:modeling_arm_kinematics} summarizes the kinematics of the robotic arm.
Section~\ref{subsec:modeling_arm_occupancy} then introduces the arm occupancy, which is the volume occupied by the arm in the environment.
Lastly, Thm.~\ref{thm:sparrows} establishes the existence of the robot's Spherical Forward Occupancy, a sphere-based reachable set representation used to construct \methodname's novel chance constraint in Section~\ref{subsec:approach_integral_evaluation}.

\subsection{Arm Kinematics}
\label{subsec:modeling_arm_kinematics}

Given a compact time interval $T \subset \R$, we define a trajectory for the robot's configuration as $q: T \to Q \subset \R^{\nq}$ and a trajectory for the velocity as $\dot{q}: T \to \R^{\nq}$.
We restate an assumption \cite[Ass. 4]{michaux2024sparrows} about the robot model:
\begin{assum}\label{assum:robot}
The robot operates in a three-dimensional workspace, denoted $\workspace \subset \R^{3}$, such that $\workspace \subset \world$ where $\world$ denotes the world frame.
There exists a reference frame called the base frame, denoted the $0$\ts{th} frame, that indicates the origin of the robot's kinematic chain.
We assume the robot's base frame coincides with the origin of the world frame.
The robot is fully actuated and composed of only revolute joints, where the $j$\ts{th} joint actuates the robot's $j$\ts{th} link.
The robot's $j$\ts{th} joint has position and velocity limits given by $\qj \in [\qlim^-, \qlim^+]$ and $\qdj \in [\dqlim^-, \dqlim^+]$ for all $t \in T$, respectively. 
The robot's input is given by $u: T \to \R^{\nq}$.
\end{assum}
We also assume that the robot's $j$\ts{th} reference frame $\{\hat{x}_j, \hat{y}_j, \hat{z}_j\}$ is attached to the robot's $j$\ts{th} revolute joint, and that $\hat{z}_j = [0, 0, 1]^\top$ points in direction of the $j$\ts{th} joint's axis of rotation.
Then $\FK_j: Q \to \R^{4 \times 4}$ maps the robot's time-dependent configuration to the pose of the $j$\ts{th} joint in the world frame\Review{:}{ such that $p_j(\q)$ and $R_j(\q)$ are the position and orientation of frame $j$ with respect to world frame $\world$, respectively.}{35-IV-1}
\Review{
\begin{equation}
\label{eq:fk_j}
    \FK_j(\q) = 
    \begin{bmatrix} R_j(\q) & p_j(\q) \\
    \zeros  & 1 \\
    \end{bmatrix},
\end{equation}
}{}{}%
\Review{where}{}{}\Review{ 
\begin{equation}\label{eq:fk_pos_j}
    \Review{p_j(\q) = \sum_{l=1}^{j} R_{l}(\q) p_{l}^{l-1},}{}{}
\end{equation}}{}{}%
\Review{and}{}{}\Review{ 
\begin{equation}\label{eq:fk_orn_j}
    \Review{R_j(\q) \coloneqq R_j^{0}(\q) = \prod_{l=1}^{j}R_{l}^{l-1}(\ql)}{}{}
\end{equation}}{}{}%
\Review{are the position and orientation of frame $j$ with respect to world frame $\world$, respectively.}{}{}

For simplicity, Assum. \ref{assum:robot} only specifies revolute joints to simplify the description of the forward kinematics; the methods described in this paper readily extend to other joint types.
We also ignore any uncertainty in the robot's dynamics, and instead assume that one could apply an inverse dynamics controller \cite{spong2005textbook} to track any trajectory of the robot perfectly.
As a result, this work focuses exclusively on modeling the kinematic behavior of the robotic arm.

\subsection{Arm Occupancy}
\label{subsec:modeling_arm_occupancy}

In this subsection, we define the \textit{forward occupancy} as the volume occupied by the arm in the workspace $\workspace$.
Let $L_j \subset \workspace \subset \R^3$ denote the volume occupied by the robot's $j$\ts{th} link with respect to the $j$\ts{th} reference frame. 
Then the forward occupancy of link $j$ is the map $\FO_j: Q \to \pow{\workspace}$ defined as
\begin{align}\label{eq:link_occupancy}
     \FO_j(\q) &= p_j(\q) \oplus R_j(\q) L_j,
\end{align}
where $p_j(\q)$ and $R_j(\q)$ specify the pose of the $j$\ts{th} joint, and $R_j(\q) L_j$ is the rotated volume of link $j$.
The volume occupied by the entire arm in the workspace is then defined by the map $\FO: Q \to \workspace$ such that
\begin{align}\label{eq:forward_occupancy}
    \FO(\q) = \bigcup_{j = 1}^{\nq} \FO_j(\q) \subset \workspace. 
\end{align}
Because the geometry of any of the robot's links may be arbitrarily complex, we restate an assumption \cite[Ass. 5]{michaux2024sparrows} that simplifies the construction of an overapproximation to the forward occupancy:
\begin{assum}\label{assum:joint_link_occupancy}
    Given a robot configuration $\q$ and any $j \in \{1,\dots, \nq\}$, there exists a ball with center $p_j(\q)$ and radius $r_j$  that overapproximates the volume occupied by the $j$\ts{th} joint in $\workspace$.
    We further assume that link volume $L_j$ is a subset of the tapered capsule formed by the convex hull of the balls overapproximating the $j$\ts{th} and $(j+1)$\ts{th} joints.
\end{assum}

Following Assum. \ref{assum:joint_link_occupancy}, we now define the ball $\Sjq$ overapproximating the volume occupied by the $j$\ts{th} joint as 
\begin{equation}\label{eq:joint_occupancy}
    \Sjq = \Btwo(p_j(\q), r_j)
\end{equation}
and the tapered capsule $TC_j(q(t))$ overapproximating the $j$\ts{th} link as
\begin{equation}
 TC_j(q(t)) = \conv\Big( \Sjq \cup \Sjpq \Big).
\end{equation}
Then, the volumes occupied by the $j$\ts{th} link \eqref{eq:link_occupancy} and the entire arm \eqref{eq:forward_occupancy} is overapproximated by
\begin{equation}\label{eq:link_capsule_occupancy}
     \FO_j(\q)  \subset TC_j(q(t))
\end{equation}
and
\begin{equation} \label{eq:forward_capsule_occupancy}
    \FO(\q) \subset \bigcup_{j = 1}^{\nq} TC_j(q(t)) \subset \workspace, 
\end{equation}
respectively.

For convenience, the notation $\FO(q(T))$ denotes the forward occupancy over an entire time interval $T$.
$\FO(q(T))$ is also called the \emph{reachable set} of the robot.
Notably, the arm is \emph{collision-free} over the time interval $T$ if $\FO(q(T))$ does not intersect with the environment.

To facilitate the exposition of our approach, we summarize the construction of the robot's safety representation in the following theorem:
\begin{thm}\label{thm:sparrows}
Given a serial manipulator with $n_q \in \N$ revolute joints and a time partition $T$ of a finite set of intervals $T_i$ (i.e., $T = \cup_{i=1}^{n_t} T_i$), the swept volume corresponding to the robot's motion over $T$ is overapproximated by a collection of $L_2$ balls in $\R^3$, which we call the \SFO ($\sfo$) \cite{michaux2024sparrows} defined as
\begin{equation}\label{eq:simple_sfo}
\sfo = \cup_{j=1}^{n_q} \cup_{i=1}^{n_t} \cup_{m=1}^{n_S} S_{j,i,m}(q(T_i; k)),
\end{equation}
where each $S_{j,i,m}(q(T_i; k))$ is an $L_2$ ball in $\mathbb{R}^3$, $n_S \in \N$ is a parameter that specifies the number of closed balls overapproximating each of the robot's links, and $k$ is a trajectory parameter that characterizes the motion of the robot over $T$.
\end{thm}

\noindent Note that one can explicitly construct an $\sfo$ that satisfies this assumption using the approach described in \cite{michaux2024sparrows}, which we summarize for convenience in Appendix \ref{app:reachability}.

\section{Detailed Description of \methodname}\label{sec:method}
This section provides a detailed overview of \methodname, a novel approach for generating risk-aware motion plans in cluttered scenes represented as radiance fields using Gaussian basis functions.
Section~\ref{subsec:approach_splatting} provides a brief overview of radiance fields and Gaussian Splatting.
Section~\ref{subsec:approach_collision} then describes how to bound the probability of collision between a ball in $\R^3$ and a learned radiance field represented by Gaussian Splats.
Section~\ref{subsec:approach_integral_evaluation} presents \Review{a closed-form}{an easily-computed expression for an}{13-V-1} upper-bound on the aforementioned probability of collision, and Section~\ref{subsec:approach_motion_planning} discusses how to leverage this result as a computationally-tractable chance constraint for online trajectory optimization.

\subsection{3D Scene Representation}
\label{subsec:approach_splatting}
This section introduces the Normalized 3D Gaussian Splatting model used to represent the environment. 
\subsubsection{Volume Rendering}
A radiance field is a 5-dimensional function $\radiance: (\x, d) \mapsto (r,g,b,\sigma)$ that maps a point $\x \in \R^3$ and viewing direction $d \in \mathbb{S}^2$ to $r,g,b$ colors and a volume density $\sigma$ \cite{volume_rendering_digest}. 
This work neglects color as it does not impact the collision probability. 
Further, the density $\sigma$ does not depend on the view direction.
As a result, we simplify the description of the radiance model by estimating the density function $\sigma~:~\R^3~\to~\R$.

We define a ray $\ray: \R \to \R^3$ by $\ray(t) = \rayorig + t\raydir$, where $\rayorig \in \R^3$ is the ray origin and $\raydir \in \mathbb{S}^2$ is the unit direction vector.
From the density function $\sigma$, we may compute the probability that a particle travels along $\ray(t)$ from $t=a$ to $t=b$ without collision using the transmittance function $\transmit_a^b$, as derived in \cite{volume_rendering_digest}: 
\begin{align} \label{eq:rendering_prob}
    \transmit_a^b[\ray] &=  \exp\left(-\int_a^b \sigma(\ray(t)) dt\right).
\end{align}
Equivalently, we may define $C_a^b[\ray]$ as the random event describing a particle colliding while traveling along $\ray$ from $a$ to $b$, where $\pr(C_a^b[\ray]) = 1 - \transmit_a^b[\ray]$ denotes the probability that a collision occurs.

\subsubsection{Normalized Gaussian Splatting}
Traditional splatting approaches represent the density function $\sigma$ using basis functions such as 3D Gaussian functions \cite{zwicker2002ewa}.
This allows the integral in \eqref{eq:rendering_prob} to be computed by transforming each Gaussian to each ray's coordinate system and then analytically marginalizing the depth dimension.
The result is a set of 2D Gaussian functions on the image plane that are queried and blended to form the image.
3DGS \cite{kerbl20233d} proposed a variation of the traditional splatting formulation that uses un-normalized 3D Gaussian functions to represent a scene.
A consequence of using un-normalized Gaussians is that the 3D to 2D transformation performed during rasterization \textit{projects} the 3D Gaussians onto the image plane rather than integrating the 3D Gaussians.
Because the 2D Gaussians are not constructed from integrating 3D Gaussians along rays, the 3D Gaussians cannot be interpreted as a basis set for $\sigma$.

In contrast, the method we introduce for collision evaluation (Section~\ref{subsec:approach_collision}) relies on using the 3D Gaussians as a basis for $\sigma$.
Hence, one of our key contributions is to re-formulate 3DGS using $n_G \in \N$ normalized 3D Gaussians such that  $\sigma$ can be used to compute the probability of collision of a particle traveling along $\varphi$.
In particular, \Review{}{}{10-V}
\begin{align}\label{eq:splatting_collision_prob}
\begin{split}
\pr(C_a^b[\ray]) &= 1 - \transmit_a^b[\ray] \\
&= 1 - \exp\left(-\int_a^b \sigma(\ray(t))dt\right) \\
&= 1 - \exp\left[\blueReview{-}\sum_{n=1}^{n_G}\frac{w_n}{\sqrt{(2\pi)^3\abs{\Sigma_n}}} \cdot \right.\\
&\quad\quad\quad\left.\int_a^b  \exp\left( -\frac{1}{2}\norm{\ray(t) - \mu_n}_{\Sigma_n^{-1}}^2 \right)dt \right],
\end{split}
\end{align}
where $w_n \in \R^+$ is a weight parameter and $G_n: \R^3 \to \R$ gives the normalized Gaussian density with mean $\mu_n \in \R^3$ and covariance matrix $\Sigma_n \in \R^{3 \times 3}$.
That is,
\begin{equation}
G_n(x)\hspace{-2pt}=\hspace{-2pt}\frac{1}{\sqrt{(2\pi)^3\det{\Sigma_n}}}\hspace{-1pt}\exp\hspace{-2pt}\left(\hspace{-2pt}\hspace{-1pt}-\hspace{-1pt}\frac{1}{2}(x\hspace{-1pt}-\hspace{-1pt}\mu_n)^T\Sigma_n^{-1}(x-\mu_n)\right)\hspace{-2pt}.
\end{equation}

Additionally, existing 3D Gaussian Splatting implementations apply a low-pass filter to the 2D Gaussian functions on the image plane by convolving the projected 2D Gaussians with an isotropic 2D Gaussian with a covariance of 0.3 pixels.
This low-pass filter reduces artifacts in the rendered images.
We omit this step from the normalized 3DGS rendering procedure because it has an effect that cannot be reasoned about in 3D.
This would undermine the validity of our collision avoidance constraint.

Instead, we apply an analogous filter in 3D Gaussians by convolving the 3D Gaussians in world coordinates with an isotropic 3D Gaussian with a small covariance of $1e-6$.
Because this operation is applied to the 3D Gaussians, it does not impact the ability of the collision constraint to use the 3D Gaussians to model $\sigma$. 
The implementation details describing the updated Gaussian Splat training process are provided in Appendix~\ref{app:normalized-3DGS}.

\subsection{Bounding the Probability of Collision}
\label{subsec:approach_collision}
\methodname enforces safety by ensuring the probability of collision between the robot and the scene is below a given risk threshold.
This subsection derives a method for bounding the probability of collision between a ball in $\R^3$ and a scene represented as a radiance field.
We make the following assumption about the Normalized 3D Gaussian Splat's representation of the scene:

\begin{assum} 
\Review{The normalized 3D Gaussian Splat is assumed to have converged such that the 3D Gaussians form a valid basis for the volume density $\sigma$.
Consequently, the transmittance, as defined in \eqref{eq:rendering_prob}, is accurately computed.}{The normalized 3D Gaussian Splat is assumed to have converged to an accurate representation of the scene, such that for all possible rays in the scene, the transmittance defined in \eqref{eq:rendering_prob} is accurately computed.}{11-V}
Furthermore, we assume that the transmittance, which represents the probability of a particle colliding while traveling along a ray, is equivalent to the probability that an infinitesimal segment of a rigid body experiences a collision when traveling along the same ray. 
\end{assum}

Without loss of generality, suppose $S =  \Btwo(0, \rho)$ is a closed $L_2$ ball centered at the origin with radius $\rho$ and whose boundary is denoted $\partial S$.
This assumption is made without loss of generality because one can apply a frame transformation to any arbitrary ball in $\R^3$ to shift its center to the origin.
Let $C(S)$ denote the random event that the ball $S$ collides with the environment.
Then, we seek to compute the probability that $C(S)$ occurs, which we denote $\pr(C(S))$.
We model this as the probability that a ray randomly cast from the center of $S$ experiences a collision before reaching $\partial S$.

Formally, let $(\Omega_\raydirvar, \mathcal{F}_\raydirvar, \mathcal{P}_\raydirvar)$ be a probability space, and let $\raydirvar:\Omega_\raydirvar \to \S^2$ be a random variable representing the direction of a randomly-cast ray originating at the center of $S$.
In particular, let $\raydirvar$ be uniformly distributed on $\S^2$ under $\mathcal{P}_\raydirvar$.
Define the random ray $\rayproc$ as:
\begin{equation}
    \rayproc(t, \omega_\raydirvar) = t\raydirvar(\omega_\raydirvar), \quad \omega_\raydirvar \in \Omega_\raydirvar.
\end{equation}
The transmittance of $\rayproc(t, \omega_\raydirvar)$ from $t=0$ to $t=\rho$ is given by
\begin{equation}\label{eq:trasnmittance-random}
    \T_0^\rho(\omega_\raydirvar) = \exp\left(-\int_0^\rho \sigma(\rayproc(t, \omega_\raydirvar))dt \right), \quad \omega_\raydirvar \in \Omega_\raydirvar.
\end{equation}
Note that $\T_0^\rho(\omega_\raydirvar)$ depends on $\rayproc(t, \omega_\raydirvar)$, which in turn depends on $\raydirvar(\omega_\raydirvar)$. 
Hence, $\T_0^\rho: \Omega_\raydirvar \to \R$ is also a random variable on $(\Omega_\raydirvar, \mathcal{F}_\raydirvar, \mathcal{P}_\raydirvar)$.
The probability the random ray collides on the interval from $t=0$ to $t=\rho$ is computed by $1 - \T_0^\rho(\omega_\raydirvar)$.
Finally, the probability that the collision risk for the randomly-cast ray exceeds a risk threshold $\alpha \in \R^+$ is given by
\begin{equation}\label{eq:sphere_collision}
    \pr(C(S)) = \pr\left(1 - \T_0^\rho(\omega_\raydirvar) \geq \alpha\right), \quad \omega_\raydirvar \in \Omega_\raydirvar.
\end{equation}

The following provides an upper bound on $\pr(C(S))$:
\begin{thm}\label{thm:collision_bound}
Consider, without loss of generality, a ball $S~=~\Btwo(0,~\rho)$ of radius $\rho$ that is centered at the origin. 
Let $\alpha \in \R^+$ denote the risk threshold defined in \eqref{eq:sphere_collision} for the probability of collision between a ray and the environment.
Then, the probability that the ball $S$ collides with the environment is bounded above by
\begin{equation}\label{eq:risk_bound}
    \pr(C(S)) \leq \frac{1}{\alpha}\hspace{-2pt}\left[1 - \exp\hspace{-2pt}\left(- \frac{1}{4\pi} \int_{S} \sigma(\x) d\x \right)\right]
\end{equation}
where the integral denotes a volume integral over the sphere $\mathcal{S}$.
\end{thm}
\noindent A proof is provided in Appendix \ref{app:collision_bound_proof}.

\subsection{Evaluating the Collision Bound}
\label{subsec:approach_integral_evaluation}
Computing the exact integral in \eqref{eq:risk_bound} is non-trivial.
While \cite[Thm~3.3]{lasserre2001integration} provides a method for computing the integral of Gaussian functions over spheres, it depends on an infinite series that is difficult to evaluate in practice.
Therefore, we derive a \Review{closed-form}{computationally attractive}{13-V-2} expression for computing an upper bound for the probability of collision in Theorem \ref{thm:collision_bound}. 
\begin{thm}\label{thm:erf_bound}
Let $S =  \Btwo(0, \rho)$ be a closed ball of radius $\rho$ that is centered at the origin.
Suppose the density function $\sigma: \R^3 \to \R$ is represented by a set of $n_G \in \N$ normalized Gaussian basis functions $\{G_n\}_{n=1}^{n_G}$ with means $\{\mu_n\}_{n=1}^{n_G}$ and covariance matrices $\{\Sigma_n\}_{n=1}^{n_G}$, each with Eigendecomposition given by $\Sigma_n = R_n\Lambda_n R_n^T$.
Finally, let the diagonal elements of $\Lambda_n$ be denoted as $\lambda_{n,1}, \lambda_{n,2}, \lambda_{n,3}$ and represent the eigenvalues of $\Sigma_n$.
Then
\begin{equation}
    \int_{S}\sigma(\x) d\x \leq \bound(S)
\end{equation}
where
\begin{align}
\bound&(S) = \sum_{n=1}^{n_G} \eta'_n w_n  \cdot \nonumber \\
& \prod_{\ell=1}^3\left[\sqrt{\frac{\pi \lambda'_{n,\ell}}{2}} \left( \erf\left(\frac{\rho - \mu'_{n,\ell}}{\sqrt{2\lambda'_{n,\ell}}}\right)
    -\erf\left(\frac{\rho + \mu'_{n,\ell}}{\sqrt{2\lambda'_{n,\ell}}}\right) \right)\right].
\end{align}

Above, $\erf$ denotes the error function and $\mu'_n$, $\lambda'_n$, and $\eta'_n$ correspond to the mean, eigenvalues, and normalization constant of the normalized Gaussian $G'_n$ obtained by rotating $G_n$ by $R^T_n$.
That is, $G'_n$ has mean $\mu'_n = R_n^T\mu_n$ and covariance $\Sigma'_n = R_n^T\Sigma_nR_n$.
Furthermore,
\begin{align}
    \pr(C(S)) &\leq \frac{1}{\alpha}\hspace{-2pt}\left[1 - \exp\hspace{-2pt}\left(- \frac{1}{4\pi} \int_{S} \sigma(\x) d\x \right)\right] \\
    &\leq \frac{1}{\alpha}\hspace{-2pt}\left[1 - \exp\hspace{-2pt}\left(- \frac{1}{4\pi} \bound(S) \right)\right].
\end{align}

Thus, we may constrain the risk of collision by enforcing
\begin{equation}\label{eq:final-risk-constraint}
    \left[1 - \exp\hspace{-2pt}\left(- \frac{1}{4\pi} \bound(S) \right)\right] < \alpha\cdot\beta
\end{equation}
for a given risk threshold $\beta \in (0,1]$. 
\end{thm}
\noindent A proof is provided in Appendix \ref{app:erf_bound_proof}. 
Note that, in practice, we set $\alpha = \beta$ for simplicity due to their mutual dependence introduced in \eqref{eq:final-risk-constraint}.

\subsection{Risk-Aware Motion Planning}
\label{subsec:approach_motion_planning}
The purpose of \methodname is to generate risk-aware motion plans in cluttered environments in a receding-horizon fashion.
Prior to planning, a normalized Gaussian Splatting representation of the scene is constructed.
At every planning iteration, the robot is given $\tplan \leq 0.5$ seconds to find a feasible trajectory by solving

\begin{align}
\underset{k \in K}{\min}& \quad \numop{cost}(k) \quad\quad\quad\quad\quad\quad\quad \splanningopt  \label{eq:splanning_opt_cost} \\
\text{s.t.}& \quad q_j(T_i; k) \subseteq [\qlim^-, \qlim^+] \quad\quad\quad \forall (i,j) \in N_t \times N_q \label{eq:pz_optpos} \\
& \quad \dot{q}_j(T_i; k) \subseteq [\dqlim^-, \dqlim^+]  \quad\quad\quad \forall (i,j) \in N_t \times N_q \label{eq:pz_optvel}\\
& \sum_{\substack{(j, m) \in \\N_q \times N_s}}\hspace{-4pt}1\hspace{-2pt} - \exp\hspace{-3pt}\left(\hspace{-3pt}- \frac{\bound(S_{j,i,m}(q_j(T_i; k))}{4\pi }  \hspace{-3pt}\right)\hspace{-3pt} <\hspace{-3pt} \alpha \cdot \beta \quad \forall i \in N_t \label{eq:pz_optcolcon}
\end{align}
where $k \in K$ is the trajectory parameter (Appendix. \ref{app:method_traj}) and $\numop{cost}(k)$ is a task-specific cost function. 
The robot's reachable set $\{S_{j,i,m}\}_{m=1}^{n_S}$ is a function of the robot's position (and hence trajectory parameter) and is computed repeatedly while numerically solving \splanningoptref{}. 
Safety is enforced using the novel collision-avoidance constraint \eqref{eq:pz_optcolcon}.
If a solution is not found, the robot executes a braking maneuver using the trajectory parameter found in the previous planning iteration.
Since \methodname's collision-avoidance constraints are differentiable, analytical constraint gradients are provided to ensure real-time motion planning.\
For simplicity, expressions for kinematics and dynamics constraints were not included in \splanningoptref.
However, \cite{michaux2023armour} and \cite{michaux2024sparrows} provide detailed explanations for including such constraints.

\begin{table*}[t!]

  \caption{Reconstruction evaluation on Replica and TUM-RGBD. 
  \Review{Each method is evaluated at 7,000 and 30,000 iterations.}{}{} 
  Standard 3DGS slightly outperforms the normalized variant. 
  However, the performance is comparable, and only Normalized 3DGS is suitable for \Review{}{risk-aware}{15-VI} motion planning. 
  \Review{The \best{best} and \ru{second best} result for each metric is annotated.}{}{}}
  \label{tab:ssim_psnr}
  \centering
  \begin{tabular}{|l|l||cccc||cccc||}
  \hline
  \multirow{2}{*}{\textbf{Dataset}}&\multirow{2}{*}{\textbf{Scene}} & \multicolumn{4}{c||}{\textbf{Unnormalized}}& \multicolumn{4}{c||}{\textbf{Normalized}} \\
  \cline{3-10}
   && PSNR & SSIM & Train Time & Render FPS & PSNR & SSIM & Train Time & Render FPS \\
   \hline\hline
\multirow{9}{*}{\textbf{Replica}}
& office0 & 44.052 & 0.986 & 1253.0 & 205.4 & 42.103 & 0.979 & 1144.1 & 174.0 \\\cline{2-10}
& office1 & 42.510 & 0.974 & 1008.9 & 218.1 & 40.457 & 0.964 & 941.1 & 180.0 \\\cline{2-10}
& office2 & 37.663 & 0.972 & 1647.6 & 174.5 & 36.055 & 0.963 & 1718.0 & 147.0 \\\cline{2-10}
& office3 & 37.533 & 0.969 & 1545.5 & 186.6 & 36.617 & 0.961 & 1636.7 & 145.4 \\\cline{2-10}
& office4 & 39.188 & 0.971 & 1488.4 & 194.2 & 38.504 & 0.968 & 1610.5 & 151.0 \\\cline{2-10}
& room0 & 37.884 & 0.971 & 1923.2 & 147.1 & 36.301 & 0.961 & 2190.4 & 108.0 \\\cline{2-10}
& room1 & 39.347 & 0.973 & 1695.3 & 165.8 & 37.513 & 0.967 & 1826.9 & 126.3 \\\cline{2-10}
& room2 & 40.280 & 0.975 & 1669.8 & 168.0 & 38.995 & 0.970 & 1635.0 & 138.1 \\\cline{2-10}
& \textbf{Average} & 39.807 & 0.974 & 1529.0 & 182.5 & 38.318 & 0.967 & 1587.8 & 146.2 \\\hline\hline
\multirow{12}{*}{\textbf{TUM-RGBD}}
& fr1/floor & 23.989 & 0.703 & 1206.2 & 190.1 & 23.626 & 0.697 & 1332.3 & 176.0 \\\cline{2-10}
& fr1/plant & 19.688 & 0.671 & 1203.0 & 194.3 & 19.189 & 0.661 & 1493.7 & 154.4 \\\cline{2-10}
& fr1/desk & 21.313 & 0.758 & 1069.2 & 259.3 & 21.018 & 0.751 & 1075.9 & 240.5 \\\cline{2-10}
& fr1/desk2 & 20.727 & 0.747 & 1146.0 & 244.1 & 20.435 & 0.743 & 1185.0 & 225.6 \\\cline{2-10}
& fr2/360\_hemi & 20.190 & 0.683 & 2479.5 & 99.9 & 19.008 & 0.664 & 3028.7 & 75.2 \\\cline{2-10}
& fr2/dishes & 23.680 & 0.819 & 1347.8 & 191.8 & 23.256 & 0.811 & 1574.3 & 148.9 \\\cline{2-10}
& fr2/flower & 22.179 & 0.763 & 1465.5 & 173.5 & 21.372 & 0.747 & 1796.5 & 127.2 \\\cline{2-10}
& fr2/coke & 21.291 & 0.773 & 1474.1 & 167.5 & 20.061 & 0.755 & 1741.7 & 126.5 \\\cline{2-10}
& fr1/room & 19.488 & 0.690 & 2368.6 & 101.3 & 19.216 & 0.691 & 2595.1 & 92.2 \\\cline{2-10}
& fr1/teddy & 19.403 & 0.652 & 1377.6 & 184.5 & 18.855 & 0.648 & 1449.2 & 169.4 \\\cline{2-10}
& fr1/360 & 20.155 & 0.717 & 1462.3 & 130.6 & 18.585 & 0.668 & 5626.5 & 33.0 \\\cline{2-10}
& \textbf{Average} & 21.100 & 0.725 & 1509.1 & 176.1 & 20.420 & 0.712 & 2081.7 & 142.6 \\\hline
  \end{tabular}\Review{}{}{30-VI.A, 16-VI.A.}
\end{table*}                            
\begin{table}
\caption{The RMSE depth error (m) is computed per evaluation image, then averaged over each image in the dataset. The standard 3DGS slightly outperforms the Normalized variant, but only Normalized 3DGS is suitable for risk-aware planning. \Review{Each method is evaluated at 7,000 and 30,000 iterations. The \textbf{best} and \underline{second best} result is annotated.}{}{}}
\label{tab:l1_depth_rmse}
\begin{tabular}{|l|l||c|c||}
\hline
& & \textbf{Normalized} & \textbf{Unormalized} \\
\hline
\multirow{9}{*}{\textbf{Replica}}
& office0 & 0.017 & 0.034 \\\cline{2-4}
& office1 & 0.013 & 0.035 \\\cline{2-4}
& office2 & 0.017 & 0.029 \\\cline{2-4}
& office3 & 0.022 & 0.036 \\\cline{2-4}
& office4 & 0.026 & 0.039 \\\cline{2-4}
& room0 & 0.023 & 0.046 \\\cline{2-4}
& room1 & 0.015 & 0.033 \\\cline{2-4}
& room2 & 0.024 & 0.034 \\\cline{2-4}
& \textbf{Average} & 0.020 & 0.036 \\\hline\hline
\multirow{12}{*}{\textbf{TUM-RGBD}}
& fr1/floor & 0.040 & 0.044 \\\cline{2-4}
& fr1/plant & 0.241 & 0.257 \\\cline{2-4}
& fr1/desk & 0.087 & 0.092 \\\cline{2-4}
& fr1/desk2 & 0.116 & 0.122 \\\cline{2-4}
& fr2/360\_hemi & 0.278 & 0.302 \\\cline{2-4}
& fr2/dishes & 0.135 & 0.154 \\\cline{2-4}
& fr2/flower & 0.311 & 0.338 \\\cline{2-4}
& fr2/coke & 0.286 & 0.307 \\\cline{2-4}
& fr1/room & 0.149 & 0.162 \\\cline{2-4}
& fr1/teddy & 0.239 & 0.241 \\\cline{2-4}
& fr1/360 & 0.155 & 0.206 \\\cline{2-4}
& \textbf{Average} & 0.185 & 0.202 \\\hline
\end{tabular}\Review{}{}{16-VI.A.}
\end{table}

\section{Experimental Results}
\label{sec:results}


This section assesses the effectiveness of \methodname by evaluating Normalized 3DGS reconstructions, the proposed constraint representation, and the proposed trajectory optimizer against baselines.

\subsection{Normalized 3DGS Evaluation}
\label{subsec:psnr_eval}
We first evaluate the ability of the normalized 3DGS to reconstruct scenes compared to the original 3DGS formulation.
We evaluate reconstruction quality on two common RGB-D datasets. First, the method was evaluated on Replica~\cite{straub2019replica}, which is a simulated dataset. 
Replica contains sequences of simulated images at 20Hz; these were decimated to 2Hz \Review{after which}{. Then,}{} 1 in every 8 retained images was excluded from training for evaluation\Review{}{, consistent with the methodology from \cite[Sec. 7.2]{kerbl20233d}}{38-VI.A}.

Next, Normalized 3DGS was evaluated on TUM-RGBD~\cite{sturm12tumrgbd}, which is a real-world dataset.
TUM-RGBD is a relatively difficult dataset for accurate 3D reconstruction due to low-resolution images and noisy depth data.
TUM-RGBD sequences were downsampled to achieve an approximate total of 250 images per sequence.
Again, 1 in every 8 images was excluded from training for evaluation.
The reported metrics were only computed on the images excluded from training.

\Review{Visual reconstruction results, measured by SSIM and PSNR~\cite{hore2010psnrssim}, are shown in Table~\ref{tab:ssim_psnr}.}
{Visual reconstruction results, measured by Structural Similarity Index Measure (SSIM) and 
Peak Signal-to-Noise Ratio (PSNR)~\cite{hore2010psnrssim}, are shown in Table~\ref{tab:ssim_psnr}.
SSIM and PSNR are widely used metrics for evaluating image reconstruction quality. 
PSNR measures the overall error between a reconstructed image and its reference by comparing the peak signal level to the noise level; it is expressed in decibels, and higher values indicate better quality. 
In contrast, SSIM assesses perceptual similarity by comparing local patterns of pixel intensities that have been normalized for luminance, contrast, and structure. 
SSIM values range from 0 to 1, with values closer to 1 indicating images that are more similar in structure. 
}{37-VI.B}%
Experiments indicate that Normalized 3DGS approaches the visual reconstruction quality of standard 3DGS
\Review{. This is a positive result because the normalization adds significant complexity to the training process and is the only method allowing risk-aware planning.}{, and with similar training and rendering speeds.
This indicates that the Normalized 3DGS representation maintains the ability of 3DGS to support rapid and high-quality view synthesis.}{
17-VI.A.}

Additionally, the geometric accuracy of each method is evaluated by measuring the difference between the rendered depth and ground-truth depth.
For each image excluded from training, a depth image is rendered and compared to the ground truth.
The Root-Mean-Square (RMS) error is calculated for each image and then averaged over all images to find the final metric.
Similar to SSIM and PSNR results, Normalized 3DGS approaches the accuracy of standard 3DGS, as shown in Table~\ref{tab:l1_depth_rmse}.

\subsection{Simulation Environment}
\begin{figure}[t]
  \centering
  \includegraphics[width=0.9\columnwidth]{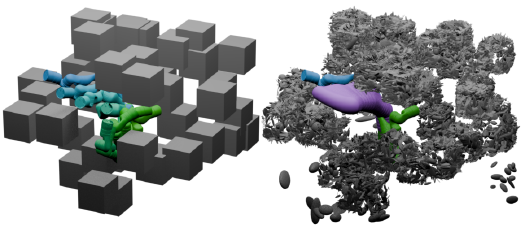} 
  \caption{
  \methodname generates safe trajectories in densely cluttered environments in simulation.
  The left panel shows discrete time steps of sequential trajectories that brings the arm from the start configuration \textcolor{StartBlue}{(blue)} safely to the goal configuration \textcolor{EndGreen}{(green)}.
  The right panel shows an intermediate planning step where the Spherical Forward Occupancy \textcolor{ReachPurple}{(purple)} avoids the obstacles represented by a Normalized 3D Gaussian Splat.
  }
  \vspace{-0.25in}
  \label{fig:40_rand_obs}
\end{figure}

\begin{figure*}[t]
    \centering
    \includegraphics[width=\textwidth]{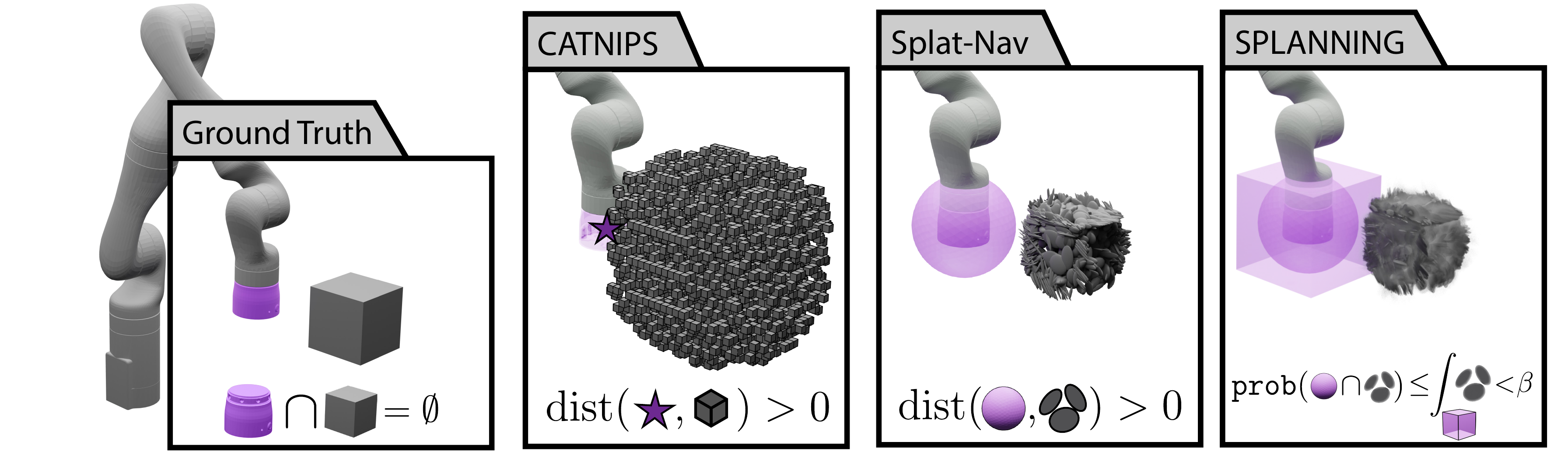}
    \caption{The collision constraint representation of \catnips, \splatnav, and \methodname are compared. \catnips transforms a NeRF into a 3D occupancy grid by relating it to a Poisson Point Process; after convolving this grid with a robot kernel, the center of the robot is checked for collision with the grid. \splatnav deterministically evaluates whether the confidence ellipsoids of each Gaussian intersect with a spherical robot. Finally, \methodname integrates a Normalized 3D Gaussian Splat over an overapproximation of the robot arm to form a risk constraint.}
    \label{fig:collision_comparison}
\end{figure*}

\begin{figure}
  \centering
  \includegraphics[width=\columnwidth]{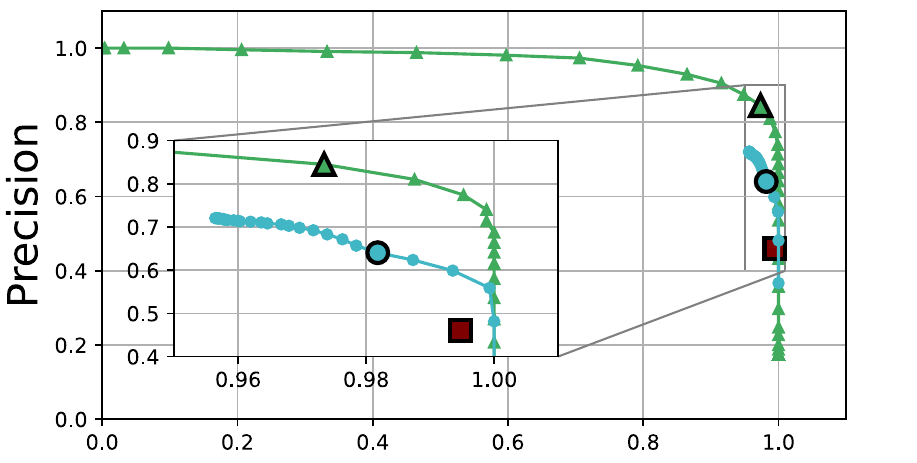}
  \includegraphics[width=\columnwidth]{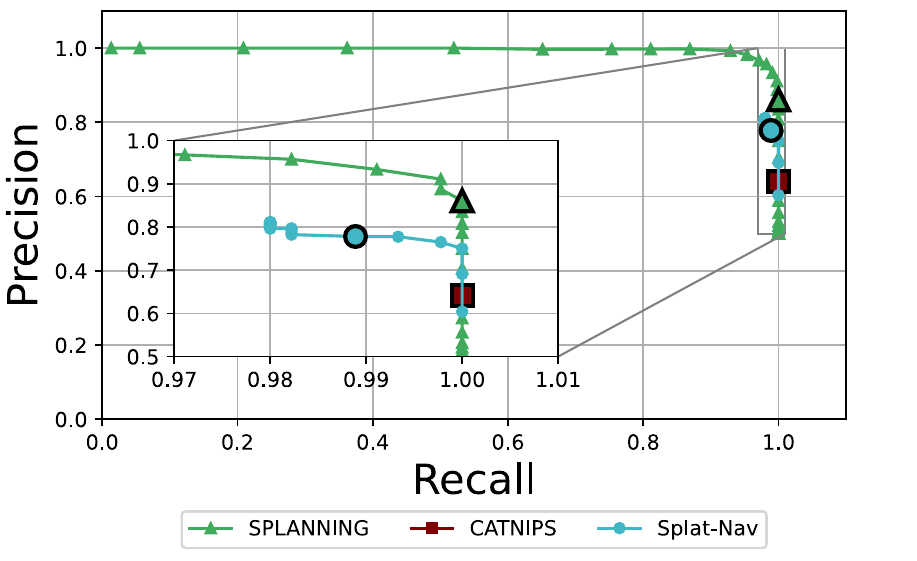}
  \caption{\Review{Treating the collision constraints from \methodname, \catnipsstar, \splatnavstar as a classifiers, where a positive indicates collision and negative indicates no collision, allows evaluating the respective precision-recall curves.}{Two Precision-Recall curves are presented. Treating the collision constraints from \methodname, \catnips, \splatnav as classifiers, where a positive indicates collision and a negative indicates no collision, the top plot shows the Precision-Recall over individual spheres while the bottom plot show Precision-Recall over configurations. Highlighted markers indicate nominal parameters $\alpha = \beta = 0.025$, $\sigma=0.99$, and $\sigma=1$ for SPLANNING, CATNIPS, and Splat-Nav respectively.}{7-VI.C, 18-VI.C}}
  \label{fig:precision_recall}
\end{figure}

The simulation environment, implemented with PyRender\footnote{\href{https://github.com/mmatl/pyrender}{https://github.com/mmatl/pyrender}}, contains cubical obstacles with 20cm sides.
The obstacles are randomly placed within the reachable space of the robot arm. 
Three sets of scenes are created with 10, 20, and 40 obstacles per scene, respectively.
For each number of obstacles, 100 random configurations are generated resulting in a total of 300 random scenes with varying degrees of clutter.
A simulated Kinova Gen3 7-DOF serial manipulator is used in the planning experiments.
The simulation environment and a corresponding 3D Gaussian Splat model are shown in Figure \ref{fig:40_rand_obs}.

\subsection{Collision Probability Evaluation}\label{subsec:collision_evals}
We first evaluate the performance of our constraint representation and compare it against existing methods that plan in radiance fields.
In particular, we compare against \catnips~\cite{chen2023catnips}, which represents the world using a Neural Radiance Field, and \splatnav~\cite{chen2024splatnav}, which represents the world using an un-normalized 3D Gaussian Splat.
Each method's collision representation is shown in Fig.~\ref{fig:collision_comparison}.
The NeRF representation for \catnips was trained using \Review{Instant-NGP~\cite{mueller2022instant}}{Nerfstudio~\cite{nerfstudio}}{18-VI.C}, while the \splatnav scene representations were built using the original Gaussian Splatting method~\cite{kerbl20233d}.
We treat each collision detection method as a classifier to determine whether a configuration will result in a collision. 
The performance of each method is measured by evaluating the precision and recall of the collision classifier~\cite{powers2008precrec}.
A True Positive indicates the correct identification of a collision, while a True Negative indicates the correct identification of free space.
 
\Review{}{We select 15 scenes, 5 for each of 10, 20, and 40 randomly-placed obstacles to sample within.}{}
In each scene, the arm configurations are classified into three categories: Unsafe (in collision), Nearly Safe ($<40$cm to the nearest obstacle), and Safe ($>40$cm to the nearest obstacle).
We randomly sample 30 configurations of the arm, which includes 10 configurations from each category to ensure an informative collection of samples.
\Review{Since the \catnips and \methodname methods are described for spherical robots, we conduct our experiments by evaluating the collision over the individual spheres that comprise the robot's spherical forward occupancy.
Further, to simplify the comparison with \catnips, we set each sphere to have the radius of the largest sphere in the SFO. 
Finally, \catnips and \splatnav are evaluated using our re-implementations of their respective constraints; as a result, we refer to their respective results as \catnipsstar and \splatnavstar.
Descriptions of the re-implementations are detailed in Appendix \ref{app:impl_details}.}
{We conduct our experiments under two conditions.
First, we compare estimated collisions to ground truth collisions over the individual spheres that comprise the robot's $\sfo$ for stationary configurations.
Second, we compare each method's estimate of whether an entire configuration is in collision against the ground truth.
When evaluating collisions, we use the radius provided by the $\sfo$ for \splatnav and \methodname, and as \catnips utilizes a fixed-size robot kernel, we fix each sphere to have the radius of the largest sphere in the $\sfo$.
Descriptions of how these baselines are implemented are detailed in Appendix~\ref{app:impl_details}.} {7-VI.C, 18-VI.C}



We calculate the Precision-Recall values at various thresholds corresponding to the allowable collision risk.
\methodname, \catnips, and \splatnav each express this maximum allowable risk differently.
For \methodname, we vary the parameters $\alpha$ and $\beta$, in the range of (0,1\Review{)}{]}{}, as discussed in Section~\ref{subsec:approach_collision}.
For \catnips, we vary the collision probability $\sigma$ \Review{and the maximum number of auxiliary particles that may exist in the robot's volume, denoted $N^{max}_{aux}$.}{in the range of (0,1] and leave all other parameters as their respective defaults from~\cite{chen2023catnips}.}{}
\Review{We refer the interested reader to~\cite{chen2023catnips} for a detailed explanation of these parameters.
In our experiments, $N_{aux}^{max}$ is set to $\{1,5,10\}$ while using $V_{aux}$ as $10^{-6}$, and $\sigma$ is varied in the range of (0,1).}{}{18-VI.C}%
Finally, \splatnav deterministically checks whether the robot collides with the \Review{99\% confidence}{1-sigma}{} ellipsoids, meaning the method does not emit a risk parameter.
Instead, we evaluate the method across several level sets of the Gaussian functions, ranging from the \Review{1-sigma}{$10^{-5}$-sigma}{} ellipsoid to the 10-sigma ellipsoid.

We report Precision-Recall curves in Fig.~\ref{fig:precision_recall}.
\Review{%
\catnipsstar achieves a high precision, indicating that most of the detected collisions were true collisions.
However, the comparatively low recall indicates that unsafe configurations were classified as safe.
\splatnavstar achieves higher recall, indicating that the method is more conservative and will result in fewer collisions.
However, the lower precision indicates the presence of false positives, which would cause a motion planner to behave overly conservatively.
The \methodname constraint achieves higher precision and recall than either \catnipsstar or \splatnavstar.
}{%
All three methods achieve high recall, indicating that all were able to classify unsafe configurations correctly.
\catnips has the lowest precision, indicating the presence of false positives which would cause a motion planner to behave overly conservatively, and has very limited control over the behavior of their constraint when $\sigma$ is varied.
\splatnav achieves higher precision and recall, with some control over the constraint behavior.
The \methodname constraint achieves the highest precision and recall.}{7-VI.C, 18-VI.C}
Further, Fig.~\ref{fig:precision_recall} shows that the thresholds $\alpha$ and $\beta$ in \methodname allow for significant control of the behavior of the constraint, ranging from highly aggressive near the left to more conservative on the right.

\subsection{Planner Performance Evaluation}
\label{subsec:planner_eval}

In this section, we quantitatively evaluate the performance of \methodname on the Kinova 7-DOF arm within a PyRender simulation environment.
In each of the 300 randomly generated scenes, a start and end pose are sampled from the free space of the scene; note that it is not guaranteed that a collision-free path exists between the start and the goal.

\def\planningcaption{Number of successes for \methodname{}, \sparrows, \armtd, \mpot, \chomp, and TrajOpt in the Kinova planning experiment with 10, 20, and 40 randomly-placed obstacles. For each set of experiments, 100 randomly-generated scenes were tested. The first number reported is the number of successes (higher is better). \crash{Red} indicates the number of failures due to collision (lower is better). The \best{most} and \ru{second most} successes are annotated.}

\begin{table*}[h!t!]
    \centering
    \caption{Number of successes for \methodname{}, \sparrows, \armtd, \mpot, \chomp, and TrajOpt in the Kinova planning experiment with 10, 20, and 40 randomly-placed obstacles. For each set of experiments, 100 randomly-generated scenes were tested. The first number reported is the number of successes (higher is better). The second number, \stuck{orange}, indicates trials that failed to reach the goal state but did not result in crashes. \crash{Red} indicates the number of failures due to collision (lower is better). The \best{most} and \ru{second most} successes are annotated.\Review{}{}{5-VI.D, 39-VI.D}}
    \begin{tabular}{|c||c||c||c||c||}
    \hline
    \multirow{2}{*}{Methods} & \multirow{2}{*}{Scene Representation} & \multicolumn{3}{c||}{\# Successes}\\\cline{3-5}
    {} & {} & 10 Obstacles & 20 Obstacles & 40 Obstacles \\
    \hline \hline
    \methodname, Risk = 0.01 & \textbf{Normalized 3DGS} & \success{29}, \stuck{ 71 }, \crash{0} & \success{6}, \stuck{ 94 }, \crash{0} & \success{0}, \stuck{ 100 }, \crash{0}\\\hline
    \methodname, Risk = 0.025 & \textbf{Normalized 3DGS} & \success{60}, \stuck{ 39 }, \crash{1} & \success{47}, \stuck{ 52 }, \crash{1} & \success{25}, \stuck{ 75 }, \crash{0}\\\hline
    \methodname, Risk = 0.0375 & \textbf{Normalized 3DGS} & \success{63}, \stuck{ 37 }, \crash{0} & \success{50}, \stuck{ 46 }, \crash{4} & \success{26}, \stuck{ 71 }, \crash{3}\\\hline
    \methodname, Risk = 0.05 & \textbf{Normalized 3DGS} & \success{\ru{69}}, \stuck{ 30 }, \crash{1} & \success{\ru{52}}, \stuck{ 43 }, \crash{5} & \success{\ru{28}}, \stuck{ 61 }, \crash{11}\\\hline
    \methodname, Risk = 0.1 & \textbf{Normalized 3DGS} & \success{67}, \stuck{ 25 }, \crash{8} & \success{35}, \stuck{ 29 }, \crash{36} & \success{25}, \stuck{ 37 }, \crash{38}\\\hline
    \sparrows & Ground-Truth & \success{\best{85}}, \stuck{ 15 }, \crash{0} & \success{\best{69}}, \stuck{ 31 }, \crash{0} & \success{\best{41}}, \stuck{ 59 }, \crash{0}\\\hline
    \armtd & Ground-Truth & \success{56}, \stuck{ 44 }, \crash{0} & \success{17}, \stuck{ 83 }, \crash{0} & \success{0}, \stuck{ 100 }, \crash{0}\\\hline
    \chomp \cite{Zucker2013chomp} & Ground-Truth & \success{30}, \stuck{ 70 }, \crash{0} & \success{9}, \stuck{ 91 }, \crash{0} & \success{4}, \stuck{ 96 }, \crash{0}\\
    \hline 
    \trajopt \cite{Schulman2014trajopt} & Ground-Truth & \success{33}, \stuck{ 0 }, \crash{67} & \success{9}, \stuck{ 0 }, \crash{91} & \success{6}, \stuck{ 0 }, \crash{94}\\
    \hline 
    \mpot \cite{le2023mpot} & Ground-Truth & \success{58}, \stuck{ 0 }, \crash{42} & \success{23}, \stuck{ 0 }, \crash{77} & \success{9}, \stuck{ 0 }, \crash{91}\\
    \hline 
    \curobo \cite{sundaralingam2023curobo} & Ground-Truth & \success{59}, \stuck{ 0 }, \crash{41} & \success{45}, \stuck{ 0 }, \crash{55} & \success{22}, \stuck{ 0 }, \crash{78}\\
    \hline
    \end{tabular}
    \label{tab:random_obstacles_0.5s}
\end{table*}
\begin{table}
    \centering

    \caption{\Review{}{The peak memory usage of SPLANNING was measured for each execution, then averaged across the 100 randomly-generated scenes for each number of obstacles.}{}}
    \Review{}{
    \begin{tabular}{|c||c|}
    \hline
    \multirow{2}{*}{Numb Obs.} & Mean Peak Memory \\
    &$\pm$ Std. Dev. (GB) \\ \hline\hline
    10 & 12.36 $\pm$ 0.98 \\\hline
    20 & 17.55 $\pm$ 1.05 \\\hline
    40 & 20.86 $\pm$ 0.98 \\\hline
    \end{tabular}}{22-VI.D}
    \label{tab:splanning_memory}
\end{table}

We compare \methodname to state-of-the-art trajectory optimization methods, including \sparrows~\cite{michaux2024sparrows}, \armtd~\cite{holmes2020armtd}, \chomp~\cite{Zucker2013chomp}, \trajopt~\cite{Schulman2014trajopt}, \mpot~\cite{le2023mpot}, and \curobo~\cite{sundaralingam2023curobo}.
\Review{All of the baseline methods have access to the ground-truth scene, while SPLANNING plans in a Normalized 3D Gaussian Splat.}{Each baseline method uses privileged geometry information from the simulation to inform collision avoidance (i.e., the exact 3D geometry of the environment is known), while SPLANNING uses simulated perception to reconstruct a Normalized 3DGS for planning.}{40-VI.D}
To evaluate whether each method produces collision-free trajectories, the simulator checks whether the robot is in collision with the ground-truth scene.
In these experiments, \methodname, \sparrows, and \armtd are limited to 0.5s to construct the plan, and each trial is run for a maximum of 150 planning horizons.
A success indicates that the robot successfully reached the goal.
Collisions are detected by checking for collisions in simulation between the robot mesh and the obstacle meshes.
Failures are also reported if the robot fails to reach the goal in the 150 planning horizons or if the optimizer fails to find a feasible plan for two consecutive planning iterations. 

For the following evaluations, the experiments involving \methodname, \sparrows, and \armtd were conducted on a system with an AMD Ryzen 5950X @ 3.4GHz and dual NVIDIA RTX A6000 GPUs.
The experiments involving CUROBO and MPOT were conducted on a system equipped with an Intel Core i7-8700K CPU @ 3.70GHz and dual NVIDIA RTX A6000 GPUs.
Baseline experiments for CHOMP and TrajOpt were performed on a machine with an Intel Core i9-12900H CPU @ 4.90GHz.
\Review{}{During all simulation experiments, only one of the two GPUs was used for trajectory optimization. In the hardware experiments described below, one GPU ran the optimizer while the other computed high-level plans in parallel. }{22-VI.D}

Table~\ref{tab:random_obstacles_0.5s} presents the number of successes and collisions achieved by each planner in the experiments.
\sparrows, which has access to the ground-truth scene, had the highest number of successes and zero collisions.
This is the expected result since \sparrows solves a similar optimization problem to \methodname but is given ground-truth scene knowledge.
In contrast, \methodname is the only method to incorporate perception.
When \methodname's $\alpha$ and $\beta$ parameters are set between $0.025$ and $0.05$, \methodname achieves more successes than all other baselines.
Closest behind \methodname are \curobo and \mpot, but both come with a significantly higher number of collisions.

The time to formulate and solve each optimization problem was measured.
For \methodname, the timings were averaged over all $\alpha,\beta$ parameters.
Further, \methodname requires a warm-up for compiled PyTorch functions; as a result, the first planning cycle is excluded from the timing results.

Timing results are summarized in Table~\ref{tab:planningtime3d7link0.5s}.
All methods remain under the 0.5s limit.
Notably, \methodname optimizes faster than \armtd for the 20 and 40 obstacle cases, despite \armtd having access to ground-truth obstacles and \methodname incorporating perception.

\begin{table}
        \centering
        \caption{Mean per-step planning time for \methodname{}, \sparrows, and \armtd in Kinova planning experiments with 10, 20, and 40 obstacles with a 0.5s time limit. For \methodname, the first planning horizon is excluded from each timing measurement as the system warms up.}
                
        \begin{tabular}{| c || c | c | c |}
            \hline
            \multirow{2}{*}{\textbf{Methods}} & \multicolumn{3}{c|}{\textbf{Mean Planning Time [s] }} \\
            \cline{2-4}
            & \textbf{10 Obstacles} & \textbf{20 Obstacles} & \textbf{40 Obstacles} \\
            \hline \hline
            \methodname{} &\Review{0.193}{0.202}{} &\Review{0.257}{0.277}{} &\Review{0.301}{0.319}{}\\
            \hline
            \sparrows{} & \Review{0.120}{0.112}{} & \Review{0.138}{0.128}{} & \Review{0.163}{0.153}{} \\
            \hline
            \armtd{} & \Review{0.162}{0.106}{} & \Review{0.289}{0.213}{} & \Review{0.372}{0.332}{} \\
            \hline
        \end{tabular}
        \label{tab:planningtime3d7link0.5s}
\end{table}

\begin{table}
    \centering
    \caption{
    Mean runtime for constraint and constraint gradient evaluation for \methodname{}, \sparrows, and \armtd in Kinova planning experiments with 10, 20, and 40 obstacles. For \methodname, the first planning constraint evaluation is excluded from each timing measurement as the system warms up.}        
    \begin{tabular}{| c || c | c | c |}
        \hline
        \multirow{2}{*}{\textbf{Methods}} & \multicolumn{3}{c|}{\textbf{Mean Constraint Evaluation Time [ms]}} \\
        \cline{2-4}
        & \textbf{10 Obstacles} & \textbf{20 Obstacles} & \textbf{40 Obstacles} \\
        \hline \hline
        \methodname{} &\Review{6.49}{6.56}{} &\Review{7.82}{7.68}{} &\Review{7.59}{8.06}{}\\
        \hline
        \sparrows{} & \Review{2.26}{2.27}{} & \Review{2.60}{2.59}{} & \Review{3.28}{3.25}{} \\
        \hline
        \armtd{} & \Review{2.82}{2.67}{} & \Review{3.52}{3.30}{} & \Review{4.10}{3.92}{} \\
        \hline
    \end{tabular}

        \label{tab:consevalruntime3d7link0.5s}
\end{table}

The time to compute each constraint and its gradient is evaluated in Table~\ref{tab:consevalruntime3d7link0.5s}.
The mean constraint evaluation for \methodname is under 10ms in all scenarios.
\methodname takes slightly longer for each constraint evaluation than \sparrows and \armtd, both of which use ground-truth representations of obstacles rather than the Normalized 3DGS used by \methodname.
However, as shown in Table~\ref{tab:planningtime3d7link0.5s}, \methodname comfortably remains under the 0.5s planning time limit.

\subsection{Real-World Demonstrations}
The planner was evaluated on a Kinova Gen3 7-DOF manipulator in a real-world setting.
For the hardware implementation, a high-level planner was added to the system.
In particular, we deploy an open-source implementation~\cite{sucan2012ompl} of bidirectional RRT~\cite{kuffner2000birrt}.
The RRT planner evaluated collisions at discrete poses by computing the spherical forward occupancy of the static arm, and evaluating the \methodname collision constraint.\Review{}{}{31-VI.E}
Additionally, to deal with the non-uniform lighting conditions arising from real-world sensor data, we augment the Normalized 3D Gaussian Splatting method to include the appearance embedding method proposed by~\cite{kulhanek2024wildgaussians}.

The hardware evaluations run the motion planner in real-time.
To enable this, four processes are run on a desktop computer with an AMD Ryzen 5950X and two NVIDIA A6000 GPUs.
First, the high-level planner takes the start and goal configuration and computes a sequence of joint waypoints that are each collision-free, as measured by the \methodname constraint.
Second, a \methodname optimizer iteratively solves the optimization problem defined in Section~\ref{subsec:approach_motion_planning} to compute a control parameter $k$.
Third, a low-level controller consumes the control parameter $k$ and uses a combination of RNEA-based inverse dynamics~\cite{luh1980RNEA} and a PD error tracking term.
Fourth, an orchestration process manages the parallel execution of the above three components.
By assuming the robot tracks the target trajectories perfectly, the \methodname process may compute the next plan as the current plan is being executed.
Similarly, the high-level planner computes the next high-level plan as the current sequence of waypoints is being tracked.
\begin{table}
\centering
\caption{\Review{}{Hardware experiments commanded a robot to sets of three high-level waypoints then back to the start. Each trial listed corresponds to one such set of motions. A trial is counted a success only if all four motions succeed; otherwise, the first failure is reported. Asterisks$^*$ indicate the number of trials where the robot gently grazed an obstacle, but motion was not impeded. RRT and Optimization Time (s) were averaged across all executions for each configuration.}{23-VI.E, 24-VI.E, 42-VI.E}}
\Review{}{
\begin{tabular}{|c||c|c|c|c||c|c|}
\hline
{\footnotesize Risk Level} & \multirow{2}{*}{\footnotesize Success} & RRT & \multirow{2}{*}{Stuck} & \multirow{2}{*}{Crash} & RRT & Opt.\\
{\footnotesize(RRT, Opt.)} & & Fail & & & {Time}& {Time}\\
\hline\hline
0.01,0.05 & 2$^{**}$ & 15 & 0 & 1 & 14.59 & 0.319 \\\hline
0.025, 0.025 & 15$^{**}$ & 3 & 0 & 0 & 16.53  & 0.329 \\\hline
0.05, 0.01 & 14 & 1 & 2 & 1 & 13.17 & 0.338\\\hline
0.1, 0.01 & 9$^*$ & 0 & 8 & 1 &12.12 & 0.351\\\hline
\end{tabular}}{}
\label{tab:hardware_results}
\end{table}

The robot was mounted in an indoor setting with two sets of shelves, as shown in Fig.~\ref{fig:opening_figure}.
For each set of shelves, three target configurations were placed in the environment.
For each set of shelves, the robot was commanded to initialize at a safe configuration, cycle between the three target configurations, and then return to the home position.
\Review{Each cycle was repeated starting with each of the three configurations it contains, meaning that three cycles were attempted per set of shelves. This results in three cycles per set of shelves, and six overall cycles.}
{Retaining cycle order, each target configuration was treated as the starting target configuration for each set of shelves, resulting in three cycles per set of shelves and six overall cycles.}{}
Finally, each cycle was repeated for three trials, resulting in a total of 18 trials.
\Review{
The $\alpha$ and $\beta$ parameters were selected to both be $0.025$, which Table~\ref{tab:random_obstacles_0.5s} demonstrates to provide a good balance between success and safety (i.e., avoiding collisions).

Of the 18 total trials, 14 resulted in success.
Of the four failures, one resulted from the arm gently grazing an obstacle, though the arm was still able to complete the trial.
One failure resulted from the high-level planner failing to find a solution.
One failure resulted from the trajectory optimizer getting stuck.
Finally, one failure resulted from a collision with the shelves.
}{%
This configuration was run with a range of risk thresholds for both the RRT high-level planner and the trajectory optimizer.
Table \ref{tab:hardware_results} provides the results.
When the RRT is run with a low risk threshold, it frequently fails to find a solution.
These failures arise both due to (a) incorrectly classifying the high-level waypoints as in collision and (b) failure to connect the start and goal configurations.
Conversely, a high RRT risk threshold combined with a low optimizer risk threshold leads to high-level plans that the trajectory optimizer is unable to track. This results in the optimizer frequently becoming stuck, shown in the final row of Table \ref{tab:hardware_results}.
When risk thresholds are chosen according to the results in Sections \ref{subsec:collision_evals} and \ref{subsec:planner_eval}, a high rate of success is achieved.

Table \ref{tab:hardware_results} also provides timing metrics for the hardware demonstrations.
The RRT timing metrics exclude cases where the RRT classifies a high-level waypoint as in-collision, in which case the planner immediately reports a failure.
Optimization times are averaged across all planning trials.
Despite these multiple processes running on the computer, real-time trajectory optimization is maintained.

}{}

\section{Conclusion}
\label{sec:conclusion}

This paper introduced a framework for evaluating collisions in a radiance field model.
The result is \methodname, a trajectory optimizer that synthesizes trajectories while constraining the probability of colliding with the environment. 
To enable the constraint formulation, we presented a normalized variant of 3D Gaussian Splatting suitable for risk-aware planning while achieving similar reconstruction quality to standard 3DGS.
The risk-aware constraint was demonstrated to outperform other radiance field methods for collision avoidance while also being efficient to compute in real-time planning.
Planner evaluations demonstrated that the trajectory optimizer achieves a high number of successes in cluttered environments.
Hardware demonstrations indicated that the planner is viable for real-world use.

There are several promising avenues for future work.
First, adding support for dynamic objects would aid in planning in real-world scenarios.
Second, the Normalized 3DGS may be improved to increase the reconstruction quality.
A key difference between Normalized 3DGS and standard 3DGS is the difference in filtering techniques; standard 3DGS applies a screen-space low-pass filter, while Normalized 3DGS applies the filter in world-space.
This adaptation allows for a probabilistic interpretation of the 3D Gaussians but is suboptimal for visual reconstruction.
Hence, future work avenues will explore other low-pass filtering techniques that improve reconstruction quality while maintaining the correctness for planning.
\Review{}{Furthermore, in instances where the Normalized 3DGS is imperfect, another future research direction will explore closed-loop, active exploration strategies for improving the reconstruction quality online.}{32-VII-1}
\Review{}{Next, in our present formulation, although $\alpha$ and $\beta$ are conceptually distinct, they are mathematically interchangeable.
Future work will investigate this issue by deriving alternate chance constraints leveraging different approximation inequalities other than Markov's inequality.}{14-VII}
Finally, incorporating a real-time simultaneous localization and mapping (SLAM) module would enable navigation of scenes that have not been previously mapped.

\printbibliography
\begin{appendices}
    \section{Trajectory Design }
\label{app:modeling_trajectory}
\methodname computes risk-aware trajectories in a receding-horizon manner by solving an optimization at each planning iteration.
Offline, we pre-specify a continuum of trajectories over a compact set $K \subset \R^{n_k}$, $n_k \in \N$. 
Then each trajectory, defined over a compact time interval $T$, is uniquely determined by a by a \textit{trajectory parameter} $k \in K$.
Note that we design $K$ for safe manipulator motion planning, but $K$ may also be designed for other tasks and robot morphologies \cite{holmes2020armtd, kousik2020bridging, kousik2019safe, liu2022refine} as long as it satisfies the following properties:

\begin{defn}[Trajectory Parameters]
\label{defn:traj_param}
For each $k \in K$, a \emph{parameterized trajectory} is an analytic function $q(\,\cdot\,; k) : T \to Q$ that satisfies the following properties:
\begin{outline}[enumerate]
\1 The parameterized trajectory starts at a specified initial condition $(q_0, \dot{q}_0)$, so that $q(0; k) = q_0$, and $\dot{q}(0; k) = \dot{q}_0$.
\1 Each parameterized trajectory brakes to a stop such that $\dot{q}(\tfin; k) = 0$.
\end{outline}
\end{defn}
\noindent \methodname{} performs real-time receding horizon planning by executing the desired trajectory computed at the previous planning iteration while constructing the next desired trajectory for the subsequent time interval.
Therefore, the first property ensures each parameterized trajectory is generated online and begins from the appropriate future initial condition of the robot.
The second property ensures that a braking maneuver is always available to bring the robot safely to a stop.

\section{Reachability Analysis}
\label{app:reachability}
In Section~\ref{app:method_traj} we discuss how to over-approximate the robot's desired position and velocity trajectories using polynomial zonotopes \cite{kochdumper2020polyzono}.
In Section~\ref{app:method_sfo} we use the over-approximated trajectories to construct an over-approximation of the forward occupancy in \eqref{eq:forward_capsule_occupancy} called the \SFO.

\subsection{Parameterized Trajectories} 
\label{app:method_traj}
\methodname enables risk-aware motion planning by constructing safety constraints that overapproximate the robot's position, velocity, and forward occupancy over continuous time intervals.
In this subsection, we briefly discuss how to overapproximate the parameterized trajectories introduced in Def. \ref{defn:traj_param}.

We start by choosing a timestep $\timestep$ that divides the compact time horizon $T \subset \R$ into $\nt \coloneqq \frac{T}{\timestep}$ time subintervals
\begin{equation}
    \label{eq:time_pz}
    \pz{T_i} =
        \left\{t \in T \mid 
            t = \tfrac{(i-1) + i}{2}\timestep + \tfrac{1}{2} \timestep \tvari,\ \tvari \in [-1,1]
        \right\},
\end{equation}
\noindent where each subinterval is indexed by the set $N_t := \{1,\ldots,\nt\}$.
Note that because intervals are a special case of polynomial zonotopes \cite{kochdumper2020polyzono}, each $\pz{T_i}$ is also a polynomial zonotope.
We can similarly represent the compact, $n_k$-dimensional trajectory parameter set $K \subset \R^{n_k}$ as a polynomial zonotope $\pz{K}$.
Then, the rules of polynomial zonotope arithmetic \cite{kochdumper2020polyzono, michaux2023armour} allow one to overapproximate the desired position and velocity trajectories by plugging $\pz{T_i}$ and $\pz{K}$ into the formulas for $\qkj$ and $\qdkj$.
We restate this result as a lemma whose proof can be found in \cite{michaux2023armour}.
\begin{lem}[Parameterized Trajectory PZs]
\label{lem:pz_desired_trajectory}
The parameterized trajectory polynomial zonotopes $\pzqji$ are overapproximative, i.e., for each $j \in \nq$ and $k\in \pz{K_j},$ 
\begin{equation}\label{eq:pz_desired_trajectory}
    \qkj \in \pzqjki \quad \forall t \in \pz{T_i}
\end{equation}
Similarly, one can define $\pzqdji$ that are also overapproximative. 
\end{lem}

\subsection{Spherical Forward Occupancy} 
\label{app:method_sfo}
This subsection briefly describes how to overapproximate the forward occupancy introduced in Section~\ref{subsec:modeling_arm_occupancy} with a collection of three-dimensional balls.
The key idea is that one can plug the polynomial zonotope trajectories \eqref{eq:pz_desired_trajectory} into a polynomial zonotope version of the robot's forward kinematics \Review{\eqref{eq:fk_j}-\eqref{eq:fk_orn_j}}{}{} to get an overapproximation of the position of each joint over the time interval $\pz{T_i}$.
Each joint position can then be overapproximated by a three-dimensional ball and added to the nominal joint ball (Assum. \ref{assum:joint_link_occupancy}) using the Minkowski sum operation.
Then by Assum. \ref{assum:joint_link_occupancy}, each link is contained in the tapered capsule $\TCjik$ of two successive joint balls over the time interval $\pz{T_i}$.
The following theorem, which originally appeared in  \cite[Theorem 10]{michaux2024sparrows}, summarizes the results of applying these steps and proves that $\TCjik$ can be overapproximated by a finite number of $\ns$ balls.

\begin{thm}[Spherical Forward Occupancy]\label{thm:sfo}
Let $\ns \in \N$ be given.
Then for each $m \in \Ns = \{1, \cdots, \ns\}$ there exists a closed ball with center $\bar{c}_{j,i,m}(k)$ and finite radius $\bar{r}_{j,i,m}(k)$ denoted
\begin{equation}
    \Sbarjimk = \Btwo\left(\bar{c}_{j,i,m}(k), \bar{r}_{j,i,m}(k)\right)
\end{equation}
such that
\begin{equation} \label{eq:pztc_bound}
    \FOjik \subset \TCjik \subset \bigcup_{m=1}^{n_s} \Sbarjimk
\end{equation}
for each  $j \in \Nq$,  $k \in \pz{K}$, and $t \in \pz{T_i}$.
\end{thm}
For convenience, we refer to the union of balls on the right-hand side of \eqref{eq:pztc_bound} as the \emph{\SFO} \Review{}{($\sfo$)}{} for the $j$\ts{th} link at the $i$\ts{th} timestep.
\begin{rem}
Them. \ref{thm:sparrows} introduced the notation in \eqref{eq:simple_sfo} to simplify the description of the \SFO in the main portion of this document.
However, we note here that $S_{j,i,m}(q(\cdot;k))$ represents the same closed ball as $\Sbarjimk$ described in Thm. \ref{thm:sfo}.
\end{rem}
    \section{\methodname Proofs}
This section re-states and provides proofs for the key theorems in the collision constraint of \methodname.

\subsection{Proof of Collision Bound}\label{app:collision_bound_proof}
\setcounter{defn}{\getrefnumber{thm:collision_bound}-1}
\begin{thm}\label{thm:app-collision_bound}
Consider, without loss of generality, a ball $S~=~\Btwo(0,~\rho)$ of radius $\rho$ that is centered at the origin. 
Let $\alpha$ denote the risk threshold defined in \eqref{eq:sphere_collision} for the probability of collision between a ray and the environment.
The probability that the ball $S$ collides with the environment is bounded above by
\begin{equation}\label{eq:app-risk_bound}
    \pr(C(S)) \leq \frac{1}{\alpha}\hspace{-2pt}\left[1 - \exp\hspace{-2pt}\left(- \frac{1}{4\pi} \int_{S} \sigma(\x) d\x \right)\right]
\end{equation}
where the integral denotes a volume integral over the sphere $\mathcal{S}$ and $\alpha \in \R^+$ is a risk threshold.
\end{thm}
\newcommand{\expcol}{\E_{\omega_\raydirvar \in \Omega_\raydirvar}[1 - \T_0^\rho(\omega_\raydirvar)]}
\newcommand{\expAcol}{\E_{\omega_\raydirvar \in \Omega_\raydirvar}&[1 - \T_0^\rho(\omega_\raydirvar)]}

\begin{proof}
Consider the probability that a sphere is in collision with the scene, denoted $\pr(C(S))$ and defined in \eqref{eq:sphere_collision} as
\begin{align*}
    \pr(C(S)) = \pr\bigg(1 - \T_0^\rho(\omega_\raydirvar) \geq \alpha\bigg), \quad\omega_\raydirvar \in \Omega_\raydirvar
\end{align*}
where $\T_0^\rho(\omega_\raydirvar)$, the probability space $(\Omega_\raydirvar, \mathcal{F}_\raydirvar, \mathcal{P}_\raydirvar)$, and random variable $\raydirvar$ are as defined in Section~\ref{subsec:approach_collision}.
From Markov's Inequality, we have that
\begin{align}\label{eq:markovs_inequality}
\begin{split}
    \pr(C(S)) &= \pr\bigg(1 - \T_0^\rho(\omega_\raydirvar) \geq \alpha\bigg)\\
     &\leq \frac{\expcol}{\alpha}.
\end{split}
\end{align}
\newcommand{\teq}{=}
\newcommand{\tcdot}{\hspace{-1pt}\cdot\hspace{-1pt}}
\newcommand{\tmin}{\hspace{-1pt}-\hspace{-1pt}}
Next, we compute the expectation
\begin{align}\label{eq:expectation-eval}
\expAcol \teq \int_{\Omega_\raydirvar}\hspace{-2pt} (1 - \T_0^\rho(\omega_\raydirvar)) d\mathcal{P}_\raydirvar \\
    &\teq 1 \tmin \int_{\Omega_\raydirvar}\hspace{-2pt} \exp\left( -\int_0^\rho \sigma(\rayproc(t, \omega_\raydirvar)) dt \right)d\mathcal{P}_\raydirvar \label{eq:expand_transmittance}\\
    &\leq 1 - \exp\left( -\int_{\Omega_\raydirvar}\hspace{-2pt}\int_0^\rho \sigma(\rayproc(t, \omega_\raydirvar)) dt d\mathcal{P}_\raydirvar\right)\label{eq:jensens}
\end{align}
The equality in \eqref{eq:expand_transmittance} results from \eqref{eq:rendering_prob}, and the inequality in \eqref{eq:jensens} results from Jensen's inequality. 

Since $\raydirvar: \Omega_\raydirvar \to \S^2$ is a random variable that maps $\omega_\raydirvar$ to ray directions $\raydir \in \S^2$, and $\raydirvar$ is uniformly distributed over $\S^2$ under $\mathcal{P}_\raydirvar$, we may push forward the measure $d\mathcal{P}_\raydirvar$ to the uniform measure on $\S^2$.
As a result, we may rewrite the outer integral in \eqref{eq:expectation-eval} as an integral over $\S^2$.
To do so, we introduce the \Review{a}{}{27-App.C-1}normalizing factor \Review{of}{}{}{}$\frac{1}{4\pi}$, corresponding to the surface area of $\S^2$.
That is,
\begin{equation}\label{eq:double_integral_sphere}
\begin{split}
    \expAcol\\
    &\leq  1 \tmin \exp\left(-\frac{1}{4\pi}\int_{\mathbb{S}^2}\hspace{-2pt}\int_0^\rho \sigma(\ray(t; \raydir)) dt d\raydir\right)
\end{split}
\end{equation}
Above, $\ray(t, \raydir) = t\raydir$.
Then, the double integral in \eqref{eq:double_integral_sphere} may be recognized as the volume integral over the ball $S$, and \eqref{eq:double_integral_sphere} is equivalent to
\begin{equation}\label{eq:volume_integral}
     \expcol \leq 1 - \exp\left(-\frac{1}{4\pi }\int_{S}\sigma(\x) d\x \right).
\end{equation}
Combining \eqref{eq:volume_integral} with \eqref{eq:markovs_inequality} results in the theorem. 
\end{proof}

\subsection{Proof of Closed-Form Bound}\label{app:erf_bound_proof}
\setcounter{defn}{\getrefnumber{thm:erf_bound}-1}
\begin{thm}\label{thm:app-erf_bound}
Let $S =  \Btwo(0, \rho)$ be a closed ball of radius $\rho$ that is centered at the origin.
Suppose the density function $\sigma: \R^3 \to \R$ is represented by a set of $n_G \in \N$ normalized Gaussian basis functions $\{G_n\}_{n=1}^{n_G}$ with means $\{\mu_n\}_{n=1}^{n_G}$ and covariance matrices $\{\Sigma_n\}_{n=1}^{n_G}$ with Eigendecompositions given by $\Sigma_n = R_n\Lambda_n R_n^T$.
Finally, let the diagonal elements of $\Lambda_n$ be denoted as $\lambda_{n,1}, \lambda_{n,2}, \lambda_{n,3}$ and represent the eigenvalues of $\Sigma_n$.
Then
\begin{equation}
    \int_{S}\sigma(\x) d\x \leq \bound(S)
\end{equation}
where
\begin{align}
\bound&(S)= \sum_{n=1}^{n_G} \eta'_n w_n  \cdot \nonumber \\
& \prod_{\ell=1}^3\left[\sqrt{\frac{\pi \lambda'_{n,\ell}}{2}} \left( \erf\left(\frac{\rho - \mu'_{n,\ell}}{\sqrt{2\lambda'_{n,\ell}}}\right)
    -\erf\left(\frac{\rho + \mu'_{n,\ell}}{\sqrt{2\lambda'_{n,\ell}}}\right) \right)\right]\hspace{-3pt}.
\end{align}
Above, $\erf$ denotes the error function and $\mu'_n$, $\lambda'_n$, and $\eta'_n$ correspond to the mean, eigenvalues, and normalization constant of the normalized Gaussian $G'_n$ obtained by rotating $G_n$ by $R^T_n$.
That is, $G'_n$ has mean $\mu'_n = R_n^T\mu_n$ and covariance $\Sigma'_n = \R_n^T\Sigma_nR_n$.
Furthermore, 
\begin{align}
    \pr(C(S)) &\leq \frac{1}{\alpha}\hspace{-2pt}\left[1 - \exp\hspace{-2pt}\left(- \frac{1}{4\pi} \int_{S} \sigma(\x) d\x \right)\right] \\
    &\leq \frac{1}{\alpha}\hspace{-2pt}\left[1 - \exp\hspace{-2pt}\left(- \frac{1}{4\pi} \bound(S) \right)\right].
\end{align}
Thus, we may constrain the risk of collision by enforcing
\begin{equation}\label{eq:app-final-risk-constraint}
    \left[1 - \exp\hspace{-2pt}\left(- \frac{1}{4\pi} \bound(S) \right)\right] < \alpha\cdot\beta
\end{equation}
for a given risk threshold $\beta \in (0,1]$. 
\end{thm}

\begin{proof}
We assume without loss of generality that the ball $S$ is at the origin of the coordinate system. 
Then,
\begin{align}
\begin{split}
    \int_{S} \sigma(\x) d\x &= \int_{S} \left(\sum_{n=1}^{n_G} w_n G_n(\x)\right) d\x \\
    &= \sum_{n=1}^{n_G} w_n \int_{S} G_n(\x) d\x.
\end{split}
\end{align}
For each Gaussian $G_n$, let $\Sigma_n = R_n\Lambda_n R_n^T$ be the Eigen decomposition of $\Sigma_n$. 
Transforming $G_n$ by the rotation matrix $R_n^T$, we obtain an axis-aligned Gaussian $G'_n$ with 
mean $\mu'_n = R_n^T\mu$ and covariance matrix $\Sigma'_n = \Lambda_n = \text{diag}\{\lambda'_1, \lambda'_2, \lambda'_3\}$.
Because rotating the ball $S$ by $R_n^T$ leaves the domain of integration unchanged, we have
\begin{align}
\begin{split}
    \int_{S} \sigma(\x) d\x 
    &= \sum_{n=1}^{n_G} w_n \int_{S} G_n(\x) d\x \\
    &= \sum_{n=1}^{n_G} w_n \int_{S} G'_n(\x) d\x.
\end{split}
\end{align}

Note that the integral of an axis-aligned Gaussian over a closed ball does not necessarily have a closed-form solution.
Therefore, we overapproximate the domain of integration by replacing the closed $L_2$ ball $S$ with the closed $L_{\infty}$ ball $S_{\infty} = B_{\infty}(0, \rho)$.

Since the covariance matrix $\Sigma'_n = \text{diag}\{\lambda'_1, \lambda'_2, \lambda'_3\}$ is diagonal, we obtain
\begin{align} \label{eq:prob_overapprox}
\begin{split}
\int_{S} \sigma(\x) d\x 
&= \sum_{n=1}^{n_G} w_n \int_{S} G'_n(\x) d\x \\
&\leq \sum_{n=1}^{n_G} w_n \int_{S_\infty} G'_n(\x) d\x \\
&=\sum_{n=1}^{n_G} \eta'_n w_n \prod_{\ell=1}^3\left[\sqrt{\frac{\pi \lambda'_n}{2}}\left( \erf\left(\frac{\rho - \mu'_{n,\ell}}{\sqrt{2\lambda'_{n,\ell}}}\right)\right.\right.  + \\
&\hspace{1cm}-\left.\left.\erf\left(\frac{\rho + \mu'_{n,\ell}}{\sqrt{2\lambda'_{n,\ell}}}\right) \right)\right]\\
&= \bound(S),
\end{split}
\end{align}
where $\erf$ denotes the error function and $\mu'_n$, $\lambda'_n$, and $\eta'_n$ correspond to the mean, eigenvalues, and normalization constant of the normalized Gaussian $G'_n$.
The result in Theorem \ref{thm:erf_bound} is now obtained by substituting \eqref{eq:prob_overapprox} into \eqref{eq:risk_bound}.    
\end{proof}

    \section{Baseline Implementation Details}\label{app:impl_details}

This section details the implementation of the baselines in the constraint comparisons of Section~\ref{subsec:collision_evals}.
\Review{Since only the collision checking mechanism was evaluated for \splatnav and \catnips, we re-implemented each method according to the details in each paper.}{}{}

\subsection{CATNIPS}

\catnips is a chance-constrained planning algorithm for robot navigation in a NeRF environment.
At a high level, \catnips relates a NeRF to a Poison Point Process (PPP) and uses this relation to create an occupancy grid called a PURR (Probabilistically Unsafe Robot Region).

\subsubsection{Intensity Grid} 
First, a NeRF representation of the PyRender scene is built using \Review{Instant-NGP~\cite{mueller2022instant}}{Nerfstudio~\cite{nerfstudio}}{}.
The space is then voxelized using a uniform scale of 150 voxels per side, as outlined in the \catnips paper \cite{chen2023catnips}\Review{.
Next, we transform the NeRF density field into a Cell Intensity Grid by discretizing the space using voxel-based representation, followed by trilinear interpolation and integration, given by {\cite[Eq. 16-17]{chen2023catnips}}.
}{%
, and the NeRF density field is transformed into a discretized Cell Intensity Grid from this voxelized representation according to {\cite[Eq. 16-17]{chen2023catnips}}.
}{18-App.D}

\subsubsection{Robot Kernel}
A robot kernel is created by performing a Minkowski sum between the robot's minimal bounding sphere and a single voxel.
This kernel is convolved with the intensity grid, resulting in the robot intensity grid.
This robot intensity grid is transformed into the PURR, as described in \cite[Section 5]{chen2023catnips}.

To evaluate collisions between the robot and the scene, we check whether the center of each sphere in the arm's SFO intersects the PURR.
Due to the convolution of the intensity grid with the robot kernel, this is equivalent to checking if the sphere intersects a discretized form of the intensity grid.

\subsection{\splatnav}
\splatnav evaluates collision in an un-normalized 3D Gaussian Splat.
Hence, splatting models of each evaluation scene were constructed using the standard 3DGS formulation \cite{kerbl20233d}.
However, to ensure a fair comparison, a modified version of 3DGS was used that includes depth supervision.

\Review{We then evaluated the collisions along each sphere using the sample-based method introduced in {\cite[Corollary 1]{chen2024splatnav}}.}
{We then evaluated the collisions along each sphere using the $K(s)$ Bisection Search introduced in \cite[Algorithm 1]{chen2024splatnav}.
We note that the provided implementation checks for collisions along linear motions with a scaling factor of $\delta_x$; to adapt this methodology to collision-checking individual spheres we set $\delta_x = 10^{-8}$, effectively setting the linear motion to a constant 0.}{7-App.D}
For convenience, this result is summarized below.

\newcommand{\rob}{\mathcal{R}}
\splatnav models the robot $\rob$ and Gaussian Splat element $G$ as ellipsoids $\mathcal{E}_\rob$ and $\mathcal{E}_G$, with means $\mu_\rob, \mu_G$ and covariances $\Sigma_\rob, \Sigma_G$ respectively.
\Review{}{They additionally model motion by parameterizing $\mathcal{E}_\rob$ as moving over a line segment, $\mu_\rob(t) = \mu_{\rob0} + t\delta_x$ for $t\in[0,1]$.}{}
Let $\Sigma_G$ be decomposed by $\Sigma_G = R\Lambda_GR^T$.
Further, denote $\Lambda_G = \diag\{\lambda_G\}$, where $\lambda_G \in \R^3$ are the eigenvalues of $\Sigma_G$.
\Review{Then, \splatnav states that $\mathcal{E}_\rob \cap \mathcal{E}_G = \emptyset$ if and only if there exists $s \in (0,1)$ such that $K(s) > 1$, where}{}{}%
\Review{\begin{equation}\label{eq:K_catnips}
    K(s) = w^T\mathbf{diag}\left(\frac{s(1-s)}{\kappa + s(\lambda_G - \kappa)}\right)w,
\end{equation}}{}{}%
\Review{$\kappa \in \R$ is the maximum eigenvalue of $\Sigma_\rob$, and $w = R^T(\mu_\rob - \mu_G)$.

For each collision check, we sampled 1000 values of $s$ uniformly on the interval $(0, 1)$.}
{Then, 
\splatnav states that $\mathcal{E}_\rob \cap \mathcal{E}_G = \emptyset$ if and only if
\begin{equation}\label{eq:splatnav_safety}
    \min_{t\in[0,1]}\max_{s\in(0,1)}K(s,t) > 1,
\end{equation}
where
\begin{equation}\label{eq:K_catnips}
    K(s,t) = w^T\mathbf{diag}\left(\frac{s(1-s)}{\kappa^2 + s(\lambda_G - \kappa^2)}\right)w,
\end{equation}
$\kappa^2 \in \R$ is the maximum eigenvalue of $\Sigma_\rob$, and $w(t) = R^T(\mu_{\rob0} + t\delta_x - \mu_G)$.}{7-App.D}

    \section{Normalized 3D Gaussian Splatting}\label{app:normalized-3DGS}

\newcommand{\rorig}{r_o}
\newcommand{\rdir}{r_d}
\let\temp\phi
\let\phi\varphi
\let\varphi\temp
\let\bar\overline
\newcommand{\loss}{\mathcal{L}}
\newcommand{\rC}{\mathcal{C}} 
\newcommand{\rD}{\mathcal{D}} 
\newcommand{\pC}{c} 
\newcommand{\tn}{\Tilde{n}} 
\newcommand{\gauss}[2]{\exp\left(-\frac{1}{2}\norm{#1}_{#2}^2\right)}
\newcommand{\Du}{\Delta u}
\newcommand{\rbar}[1]{\left.#1\right|}
\newcommand{\part}[2]{\frac{\partial #1}{\partial #2}}

This appendix provides a detailed derivation of the rasterization process of a normalized 3D Gaussian Splat and the computation of its derivative with respect to the splatting parameters.
In order to remain consistent with the splatting literature, the notation in this section is self-contained and distinct from the notation used in the rest of the document.
In practice, the reference implementation of the original 3D Gaussian Splatting method \cite{kerbl20233d} was modified to include normalization in both the forward and backward pass.

\subsection{Radiance Field Preliminaries}
First, we re-state the definition of a radiance field.
A radiance field models the image formation process by approximating a five-dimensional function that maps from a point and viewing directions to the differential opacity $\sigma: \R^3 \to \R$ and the color $\pC: \R^3 \times S^2 \to \R^3$.
Intuitively, $\sigma$ represents the likelihood that a particle traveling along a ray will stop when it collides with an individual point in space.
Under a radiance field model, an image is rendered by considering separately the ray that intersects each pixel.
Consider a ray $r(s) : \R \to \R^3$.
We define an additional quantity, the \textit{transmittance}, which defines the probability that a particle traveling along ray $r$ from $s=a$ to $s=b$ will experience a collision. 
This transmittance, denoted as $\T_a^b[r]$, is computed by

\begin{equation}\label{eq:app-transmittance}
    \T_a^b[r] = \exp\left(-\int_a^b \sigma(r(s))ds\right)
\end{equation}
Equivalently, $\T_a^b[r]$ may be interpreted by considering a light source aligned with ray $r$ and located at $s=a$.
Then $\T_a^b[r]$ is the proportion of light that remains by point $s=b$.

From the transmittance, the expected color of the pixel corresponding to ray $r$ can be computed as
\begin{equation}\label{eq:rendering-equation}
    \rC[r] = \int_{z_n}^{z_f} \pC(r(s), \rdir)\sigma(r(s)) T_{z_n}^s[r]ds
\end{equation}
where $z_n, z_f$ are the near and far clipping planes. 
The expected depth can be analogously computed as
\begin{equation}\label{eq:depth-rendering-equation}
    \rD[r] = \int_{z_n}^{z_f} s \cdot \sigma(r(s)) T_{z_n}^s[r]ds
\end{equation}

\subsection{Summary of Splatting Methods}
This section summarizes splatting methods, while the rest of the Appendix provides a detailed derivation of Normalized 3DGS.
We refer the interested reader to \cite{zwicker2002ewa} for a more comprehensive treatment of splatting-based rendering.
In a splatting framework, 3D Gaussian functions are used as a basis set to approximate $\sigma$ and $\pC$.
We call each Gaussian function a 3D reconstruction kernel $\kappa(x)$ in world coordinates.
Then, each 3D function $\kappa(x)$ is transformed into a coordinate system aligned with each camera ray.
This yields a 3D reconstruction kernel $\kappa'(x)$ in ray coordinates.
Next, each kernel is integrated along the depth dimension of the ray to yield a 2D image-space reconstruction kernel $q(x)$.
Finally, the 2D kernels are queried at the center of each pixel and blended to compute the final pixel color.

\subsection{Representation}

Let $n_G$ 3D Gaussian density functions be reconstruction kernels to represent the scene.
Let each function parameterized by mean $\mu_k$ and covariance $\Sigma_k$, both expressed in world coordinates.
Denote $S_k$ = $\Sigma_k^{-1}$. 
Then, the 3D reconstruction kernel is given by the 3D Gaussian density
\begin{equation}\label{eq:gaussian-density}
    G_k(x) = \frac{1}{\sqrt{(2\pi)^3\det{\Sigma_k}}}\gauss{x - \mu_k}{S_k}
\end{equation}
where the represented norm is the Mahalanobis distance.
For notational simplicity, we denote
\begin{equation}\label{eq:normalizing_const}
    n_k = \frac{1}{\sqrt{(2\pi)^3\det{\Sigma_k}}}
\end{equation}
as the normalizing constant of the Gaussian.
Finally, the differential opacity $\sigma$ at a point $x$ is given by
\begin{align}\label{eq:gaussian-sigma-basis}
\begin{split}
    \sigma(x) &= \sum_{k=1}^{n_G} w_k G_k(x) \\
    &= w_k n_k \gauss{x - \mu_k}{S_k}
\end{split}
\end{align}
where $w_k \in \R^+$ are positive weights.

Additionally, assume without loss of generality that the Gaussian carries with it a viewpoint-independent color $\pC_k \in \R^3$.
This assumption is made without loss of generality because, in practice, a viewpoint-dependent color may be converted to a constant color once a viewpoint is selected.
Additionally, we assume the covariance $\Sigma_k$ is provided.
In implementation, $\Sigma_k$ is parameterized by its Eigendecomposition; this detail is omitted for brevity.

In summary, the parameters of each Gaussian $G_k$ in the scene representation are the mean of the Gaussian $\mu_k \in \R^3$; the covariance of the Gaussian $\Sigma_k \in \R^{3 \times 3}$; the color of each Gaussian $\pC_k \in \R^3$; and the weight of each Gaussian $w_k \in \R^+$.

\subsection{Integration of an Individual Gaussian}\label{sec:integrating-gaussians}

During the forward pass, each Gaussian is transformed into a coordinate frame aligned with each camera ray, which reduces the integration problem to a set of Gaussian marginalizations.

This section closely follows Section 6.2 of \cite{zwicker2002ewa}, with slight modifications for coordinate systems and notation.
Further, \cite{ye2023gsplatmath} provides a similar derivation for the more standard un-normalized 3D Gaussian Splatting.
We consider the integration of an individual reconstruction kernel $\kappa_k(x) = G_k(x)$ along an individual camera ray.
Let $G_k$ have mean $\mu_k$ and covariance $\Sigma_k$ expressed in world coordinates.
Denote the homogenous $SE(3)$ transformation from world coordinates to camera coordinates as
\begin{equation}
    T_{CW} = \begin{bmatrix}
        R_{CW} & d_{CW} \\
        0 & 1
    \end{bmatrix},
\end{equation}
where $R_{CW} \in SO(3)$ is a rotation matrix and $d_{CW} \in \R^3$ is a translation vector.

Consider transforming a point $x \in \R^3$ to image coordinates.
First, we transform it to camera coordinates by $t = \phi(x) = R_{CW}x + d_{CW}$, where $t \in \R^3$.
Denote the mean in camera coordinates as $\mu_k^t = \phi(\mu_k)$.
Then, points are transformed to a ray coordinate system via a projective transformation. 
We use the standard projection matrix from camera to clip space, defined as
\begin{equation}
    P = \begin{bmatrix}
        2f_x/W & 0 & 0 & 0\\
        0 & 2f_y/H & 0 & 0\\
        0 & 0 & \frac{z_f+z_n}{z_f-z_n} & -2\frac{z_fz_n}{z_f-z_n}\\
        0 & 0 & 1 & 0
    \end{bmatrix}.
\end{equation}
We simultaneously convert the first two components to pixel units via the camera's focal lengths $f_x$ and $f_y$, image size $(W, H)$, and principal point $(c_x, c_y)$: 

\begin{align}
    \varphi(t) &= 
        \begin{bmatrix}
            \frac{W P\cdot t_x}{2P\cdot t_w} + 0.5 + c_x \\
            \frac{H P\cdot t_y}{2P\cdot t_w} + 0.5 + c_y \\
            \norm{t}
        \end{bmatrix}\\
        &= \begin{bmatrix}
        f_xt_x/t_z + 0.5 + c_x \\
        f_yt_y/t_z + 0.5 + c_y \\
        \norm{t}
    \end{bmatrix}.\label{eq:image-coordinates}
\end{align}
Since the transformation \eqref{eq:image-coordinates} is not affine, it is approximated with a first-order Taylor approximation around $\mu^t$:
\begin{equation}
    \varphi_{\mu_k^t}(t) = \varphi(\mu_k^t) + J_k(t - \mu_k^t)
\end{equation}
where $J_k$ is the Jacobian of the transformation in \eqref{eq:image-coordinates}. In particular,
\begin{equation}\label{eq:J}
    J_k = \begin{bmatrix}
        f_x/\mu^t_{k,z} & 0 & -f_x\mu^t_{k,x}/{\mu^t_{k,z}}^2 \\
        0 & f_y/\mu^t_{k,z} & -f_y\mu^t_{k,y}/{\mu^t_{k,z}}^2 \\
        \mu^t_{k,x}/\norm{t_k}& \mu^t_{k,y}/\norm{\mu^t_{k}} & \mu^t_{k,z}/\norm{\mu^t_k}
    \end{bmatrix}
\end{equation}

Applying these transformations results in a ray-space reconstruction kernel given by
\begin{align}\label{eq:ray-space-kernel}
    \kappa'(u) &= \det{R_{CW}}\det{J_k}G_k'(u) \\
    \mu_k' &= \varphi(\phi(\mu_k)) \\
    \Sigma_k' &= J_kR_{CW}\Sigma R^\top _{CW}J_k^\top \label{eq:sigmaprime}
\end{align}
where $G_k'$ is a Gaussian density with mean $\mu_k'$ and covariance $\Sigma_k'$.
We refer the interested reader to \cite[Section 6.2.2]{zwicker2002ewa} for a detailed justification of the above.
Note that $R_{CW}$ is a rotation matrix, so its determinant $\det{R_{CW}} =1$ will be excluded moving forward. 

Finally, the image-space 2D reconstruction kernel is computed by marginalizing the third dimension of $\kappa'$.
As demonstrated in \cite[Section 6.2.3]{zwicker2002ewa}, this is equivalent to
\begin{equation}\label{eq:2d-reconstruction-kernel}
    q_k(\hat{u}) = \det{J_k}\hat{G}_k'(\hat{u})
\end{equation}
where $\hat{u} \in \R^2$ indicates a point on the image plane in pixel coordinates, and $\hat{G}'$ is a 2D Gaussian function with mean $\hat{\mu}_k'$ and covariance $\hat{\Sigma}'_k$ constructed by the first two rows (and columns) of the corresponding parameters of $G'_k$. We will denote the normalizing constant of $\hat{G}_k$ as $\hat{n}_k$.

\subsection{Compositing the Integrated Gaussians}
The per-Gaussian integration yields 2D reconstruction kernels which are then queried and blended onto the image plane to complete the rasterization.
In this description, we omit implementation-specific details of rasterization, including how the pixels are split into tiles, which Gaussians are included for rasterization in which tiles, and how the Gaussians are sorted.
Instead, we focus only on mathematically relevant aspects of the process, and assume without loss of generality that every Gaussian is included for rasterization in every pixel.

With this assumption, we may write a rasterization form of \eqref{eq:rendering-equation} to compute the color of a pixel with 2D coordinates $\hat{u}$
\begin{equation}\label{eq:rasterization}
    \rC(\hat{u}) = \sum_{k=1}^{n_G} w_k \pC_k q_k(\hat{u}) \prod_{j=1}^{k-1}\left(1 - w_jq_j(\hat{u}))\right).
\end{equation}
Zwicker et. al. \cite{zwicker2002ewa} provides a more thorough derivation of the above.
We may recognize the product on the right-hand-side as the transmittance $\T_k(\hat{u})$ and define $\alpha_k(\hat{u}) = w_kq_k(\hat{u})$ to simplify \eqref{eq:rasterization} to 
\begin{equation}\label{eq:rasterization-simple}
    \rC(\hat{u}) = \sum_{k=1}^{n_G}\alpha_k(\hat{u}) \pC_k \T_k(\hat{u}).
\end{equation}
In practice, the above is implemented by iterating over the Gaussian functions front-to-back once per pixel, and keeping running track of the product on the right-hand-side.
The depth is analogously rendered by
\begin{equation}\label{eq:depth-rasterization}
    \rD(\hat{u}) = \sum_{k=1}^{n_G}\alpha_k(\hat{u}) \mu^t_{k,z} \T_k(\hat{u})
\end{equation}
where $\mu^t_{k,z}$ denotes the $z$ component of the $k^{th}$ mean in the camera's frame.

\subsection{Backward Pass}
In this section, we derive the gradient of the rendered color $\rC(\hat{u})$ and depth $\rD(\hat{u})$ of the pixel $\hat{u}$ with respect to the parameters of the Gaussian Splat.
We derive the gradient for a single pixel $\hat{u} \in \R^2$; in practice, the total gradient for each Gaussian parameter results from the summation of the below derivation over all the pixels.

We may write the contribution of the $k^{\text{th}}$ Gaussian to $\rC(\hat{u})$ and $\rD(\hat{u})$ as
\begin{align}\label{eq:start-of-backward}
    \rC_k(\hat{u}) = \alpha_k(\hat{u})\pC_k\T_k(\hat{u})\\
    \rD_k(\hat{u}) = \alpha_k(\hat{u})\mu^t_{k,z}\T_k(\hat{u})
\end{align}
Next, \eqref{eq:rasterization-simple} and \eqref{eq:depth-rasterization} give that
\begin{equation}
    \part{\rC(\hat{u})}{\rC_k(\hat{u})} = \part{\rD(\hat{u})}{\rD_k(\hat{u})} = 1.
\end{equation}
Thus, we generally consider the impact of a single Gaussian on the rendered color and depth.
The dependence of variables on $\hat{u}$ will be omitted for notational simplicity.
Finally, we assume that an arbitrary loss function $\L$ is used in training.
Further, we assume that  the backward pass of the optimization provides the gradients of the loss function with respect to the rendered depth $\rD_k$ and the rendered color $\rC_k$.
We denote these input gradients as
\begin{equation*}
    \part{\L}{\rD_k}, \part{\L}{\rC_k}
\end{equation*}
respectively.
The remainder of this document specifies how to propagate these given gradients to the parameters of the Gaussian Splat.

\subsection{Gradient of the Render Process}
We begin at the end of the forward pass by differentiating \eqref{eq:rasterization-simple}, in which the integrated 2D Gaussians are composited to form pixels.
The input to the forward render pass are the 2D means $\hat{\mu}'_k$, inverses of the 2D covariances $\hat{S}'_k$, RGB colors $c_k$, normalizing constants \Review{$n_k$}{$\hat{n}_k$}{}, and the determinant of the projection operation $\det{J_k}$. For convenience, denote
\begin{equation}
    \tn_k = \det{J_k}\hat{n}_k.
\end{equation}
The backwards render pass propagates gradients with respect to $\hat{\mu}'_k$, $\hat{S}'_k$, $c_k$, and $\tn_k$.

In implementation, we iterate over the Gaussians back-to-front.
During the forward pass, we store the final transmittance $\T_{n_G}$ for each pixel such that during the backward pass, we may iteratively compute 
\begin{equation}
    \T_{k-1} = \frac{\T_{k}}{1 - \alpha_{k-1}}. 
\end{equation}

Starting with the color parameters $c_k$, we find
\begin{align}\label{eq:color-grad}
    \part{\rC_k}{\pC_k} &= \alpha_k \T_k.\\
    \part{\L}{\pC_k} &= \alpha_k \T_k \cdot \part{\L}{\rC_k}.
\end{align}
This completes the gradient with respect to the color parameter $c_k$.

Similarly, we may differentiate \eqref{eq:depth-rasterization} with respect to the input depth of the Gaussian, $\mu^t_{k,z}$:
\begin{align}\label{eq:depth-grad}
    \part{\rD_k}{\mu^t_{k,z}} &= \alpha_k \T_k\\
    \rbar{\part{\L}{\mu^t_{k,z}}}_{\rD_k} &= \alpha_k \T_k \cdot \part{L}{\rD_k},\end{align}
where the bottom term denotes the contribution to the gradient due to the rendered depth.
We will put this term aside and revisit it later in the backward pass. 

Next, we consider the derivative of the rendered color with respect to $\alpha_k$.
\begin{align}\label{eq:dck_dalpha}
    \part{\rC_k}{\alpha_k} &= \pC_k \T_k.
\end{align}

We additionally must consider the impact of $\alpha_k$ on all future colors
\Review{
, which we denote
}{
. Consider that the rendered color of a pixel can be written as
\begin{equation}
    \rC = \sum_{j=1}^k \rC_k + \bar{\rC}_k
\end{equation}
where $\bar{\rC}_k$ denotes the contributions to $\rC$ of all Gaussians after the $k^{th}$ Gaussian.
In particular,
}
{}
\begin{align}\label{eq:barc_k}
    \bar{\rC}_k = \sum_{j = k+1}^{n_G} c_j \alpha_j \T_j.
\end{align}
Differentiating with respect to $\alpha_k$ gives
\begin{align}
    \part{\bar{\rC}_k}{\alpha_k} &= -\frac{\bar{\rC}_k}{1-\alpha_k} \\
    \rbar{\part{\loss}{\alpha_k}}_{\rC} &= \part{\L}{\rC} \left(\pC_k\T_k - \frac{\bar{\rC}_k}{1-\alpha_k} \right)
\end{align}
Note that $\bar{\rC}_k$ may also be computed iteratively backwards beginning with $\bar{\rC}_{n_G} = 0$. Analogously, 
\begin{align}\label{eq:dDk_dalpha}
    \bar{\rD}_k &= \sum_{j=k+1}^{n_G} \mu^t_{j,z}\alpha_j\T_j \\
    \rbar{\part{\loss}{\alpha_k}}_{\rD} &= \part{\L}{\rD_k} \left(\mu^t_{k,z}\T_k - \frac{\bar{\rD}_k}{1-\alpha_k} \right)
\end{align}
This allows
\begin{equation}\label{eq:dL_dalphak}
    \part{\L}{\alpha_k} = \rbar{\part{\L}{\alpha_k}}_{\rC} + \rbar{\part{\L}{\alpha_k}}_{\rD}
\end{equation}

We now continue to backpropagate through $\alpha_k$. Recall that
\begin{equation}\label{eq:alpha-reminder}
    \alpha_k = w_k \tn_k\gauss{\hat{u}-\hat{\mu}'_k}{\hat{S}'_k}.
\end{equation}
Since it will be frequently used, denote
\begin{equation}
    \rho_k := \gauss{\hat{u}-\hat{\mu}'_k}{\hat{S}'_k}.
\end{equation}
From \eqref{eq:alpha-reminder}, we find
\begin{equation}
    \part{\alpha_k}{w_k} = \tn_k \rho_k,\quad
    \part{\alpha_k}{\tn_k} = w_k \rho_k,\quad
    \part{\alpha_k}{\rho_k} = w_k \tn_k,
\end{equation}
which gives
\begin{align}
    \part{\L}{w_k} &= \part{\L}{\alpha_k}\tn_k \rho_k \label{eq:dL_dwk} \\
    \part{\L}{\tn_k} &= \part{\L}{\alpha_k}w_k \rho_k \label{eq:dL_dtnk} \\
    \part{\L}{\rho_k} &= \part{\L}{\alpha_k}w_k \tn_k \label{eq:dL_drhok}.
\end{align}
The second equation \eqref{eq:dL_dtnk} completes the backward pass of the render process with respect to the input $\tn_k$.

We now continue to backpropagate through $\rho_k$.
Denote 
\begin{equation}
    \Du_k = \hat{u} - \hat{\mu}'_k.
\end{equation}
Then we may compute the partial derivative of $\rho_k$ with respect to the 2D mean $\hat{\mu}'_k$ via
\begin{equation}
    \part{\rho_k}{\Du_k} = -\rho_k\hat{S}'\Du_k
\end{equation}
and
\begin{align}\label{eq:dL_dmuhatprime}
\begin{split}
    \part{\L}{\hat{\mu}_k'} &= \part{\L}{\alpha_k}\cdot\part{\alpha_k}{\rho_k}\cdot\part{\rho_k}{\Du_k}\cdot\part{\Du_k}{\hat{\mu}'_k} \\
    &= -\part{\L}{\alpha_k}\cdot\part{\alpha_k}{\rho_k}\cdot\part{\rho_k}{\Du_k}_.
\end{split}
\end{align}
This completes the gradient of the loss with respect to the 2D mean.

We may also compute the partial derivative of $\rho_k$ with respect to the elements of $\hat{S}_k'$, the inverse of the 2D covariance, via
\begin{equation}
    \part{\rho_k}{\hat{S}'_k} = \frac{\rho_k}{2}\Du_k^\top \Du_k,
\end{equation}
which allows
\begin{equation}\label{eq:dL_dconic}
    \part{\L}{\hat{S}_k'} = \part{\L}{\alpha_k}\cdot\part{\alpha_k}{\rho_k}\cdot\part{\rho_k}{\hat{S}_k\Review{}{'}{}}.
\end{equation}
This completes the gradient of the rendered color with respect to the inverse 2D covariance.

We have now computed the gradients of $\L$ with respect to $\hat{\mu}'_k$, $\hat{S}'_k$, $c_k$, and $\tn_k$, which completes the backward pass of the rendering function.

\subsection{Gradients of the Gaussian Integration}
The second step of the backward pass is to propagate the gradients through the coordinate transformations and marginalization described in Section~\ref{sec:integrating-gaussians}.
We differentiate the outputs of the integration step, which are $\hat{\mu}'_k$, $\hat{S}'_k$, and $\tn_k$.

\paragraph{Derivatives of $\tn_k$}
We begin by considering the derivatives of $\tn_k$. Recall that 
\begin{equation}\label{eq:tnk-reminder}
    \tn_k = \frac{\det{J_k}}{2\pi\sqrt{\det{\hat{\Sigma}'_k}}}.
\end{equation}
We start by considering the derivative of $\tn_k$ with respect to the elements of $J_k$. Recall that $J_k$ appears not just in the numerator of \eqref{eq:tnk-reminder}, but also in the definition of $\hat{\Sigma}'_k$.
First, from \cite[Eq. 49]{matrixcookbook}, we have the derivative of $\tn_k$ with respect to $J_k$ due to the inclusion of $\det{J_k}$ is
\begin{equation}\label{eq:dtnk_djkdetj}
    \rbar{\part{\tn_k}{J_k}}_{\det{J_k}} = \Review{\frac{\det{J_k}J_k^{-T}}{\sqrt{2\pi\det{\hat{\Sigma}'_k}}}}{\frac{\det{J_k}J_k^{-T}}{2\pi\sqrt{\det{\hat{\Sigma}'_k}}}}{}.
\end{equation}
We now pause dealing with $J_k$, and turn our attention to the $\det{\Sigma'_k}$ term in the denominator of \eqref{eq:tnk-reminder}. 
By the same identity from \cite[49]{matrixcookbook},
\begin{equation}
    \part{\tn_k}{\hat{\Sigma}'_k} =\Review{}{-}{}\frac{\det{J_k}\left(\hat{\Sigma}'_k\right)^{-T}}{4\pi\sqrt{\det{\hat{\Sigma'_k}}}}.
\end{equation}
Continuing,
\begin{align}
    \part{\tn_k}{\Sigma'_k} &= \begin{bmatrix}
         \part{\tn_k}{\hat{\Sigma}'_k} & 0 \\
         0 & 0
    \end{bmatrix}\\
    \rbar{\part{\L}{\Sigma'_k}}_{\tn_k} &= \part{\L}{\tn_k}\cdot\part{\tn_k}{\Sigma'_k}
\end{align}
where the second term denotes the contribution to the partial derivative by $\tn_k$.

\paragraph{Derivatives of $\hat{S}'_k$}
Given the derivative $\frac{\partial \L}{\partial \hat{S}'_k}$ computed in \eqref{eq:dL_dconic}, we may repeatedly apply \cite[Eq.~60]{matrixcookbook} to find the partial derivatives with respect to $\hat{\Sigma}'_k = \left(\hat{S}'_k\right)^{-1}$.
The results are omitted for brevity.

From this, it is trivial to apply the chain rule element-wise to obtain the partial derivatives of $\L$ with respect to $\hat{\Sigma}'_k$ due to $\hat{S'}_k$, which we denote
\begin{equation*}
    \rbar{\part{\L}{\hat{\Sigma}'_k}}_{\hat{S}'_k}
\end{equation*}
Continuing, we observe
\begin{align}
    \rbar{\part{\L}{\Sigma'_k}}_{\hat{S}'_k} &= \begin{bmatrix}
         \rbar{\part{\L}{\hat{\Sigma}'_k}}_{\hat{S}'_k} & 0 \\
         0 & 0
    \end{bmatrix}.
\end{align}
Finally, we arrive at the derivative of the loss $\L$ with respect to the transformed covariance $\Sigma'_k$
\begin{equation}
    \part{\L}{\Sigma'_k} = \rbar{\part{\L}{\Sigma'_k}}_{\hat{S}'_k} + \rbar{\part{\L}{\Sigma'_k}}_{\tn_k}
\end{equation}
We conclude the derivatives with respect to $\Sigma_k$ by transforming this back to the source coordinate system by
\begin{equation}
    \part{\L}{\Sigma_k} = R_{CW}^\top J_k^\top \part{\L}{\Sigma'_k}J_kR_{CW}.
\end{equation}

Now, we continue to backpropagate the gradients with respect to $J_k$. Denote $T_k = J_kR_{CW}$. Then
\begin{align}
    \part{\L}{T_k} &= \left(\part{\L}{\Sigma'_k}^\top + \part{\L}{\Sigma'_k}\right)T_k\Sigma_k \label{eq:dck_dtk}\\
    \part{T_k}{J_k} &= R_{CW}.\label{eq:dtk_djk}
\end{align}
This brings us to
\begin{equation}
    \part{\L}{J_k} = \part{\L}{\tn_k}\rbar{\part{\tn_k}{J_k}}_{\det{J_k}} + \part{\L}{T_k}\part{T_k}{J_k}
\end{equation}
where the first term is given by \eqref{eq:dL_dtnk} and \eqref{eq:dtnk_djkdetj}, and the second is given by \eqref{eq:dck_dtk} and \eqref{eq:dtk_djk}.
$J_k$ depends on $\mu^t_k$, so we continue to backpropagate.
\begin{align}
    \part{\L}{\mu^t_k} &= \begin{bmatrix}\sum_{i=1}^3\sum_{j=1}^3 \part{\L}{J_{k,(i,j)}}\part{J_{k,(i,j)}}{\mu^t_{k,n}}\end{bmatrix}^\top_{n=\{1,2,3\}}\label{eq:dL_dmukt} \\
    &\quad + \begin{bmatrix} 0 & 0 & \rbar{\part{\L}{\mu^t_{k,z}}}_{\rD_k}\end{bmatrix}^\top\nonumber,
\end{align}
where $J_{k, (i,j)}$ denotes the $(i,j)$ entry of $J_k$. In practice, each term in \eqref{eq:dL_dmukt} was computed symbolically, and the result is omitted here for brevity.
Finally, we arrive at the gradient of $\L$ with respect to $\mu_k$ due to the contribution from $J_k$ as
\begin{equation}
    \rbar{\part{\L}{\mu_k}}_{J_k} = R_{CW}^\top\part{\L}{\mu^t_k}
\end{equation}

\paragraph{Gradient of of $\hat{\mu}_k'$}
The final term to differentiate is $\hat{\mu}_k'$.
Immediately, we find that
\begin{align}
    \part{\hat{\mu}_k'}{\mu_k} &= \part{\hat{\mu}'_k}{\mu'_k}\cdot  \part{\mu_k'}{\mu^t_k} \cdot \part{\mu^t_k}{\mu_k} \\
    &= \begin{bmatrix}
        1 & 0 & 0 \\
        0 & 1 & 0
    \end{bmatrix} \cdot J_k \cdot R_{CW}.
\end{align}
This may be used to compute
\begin{equation}
    \rbar{\part{\L}{\mu_k}}_{\hat{\mu}'_k} = \part{\L}{\hat{\mu}'_k}\cdot \part{\hat{\mu}_k'}{\mu_k}.
\end{equation}
This gives the final gradient,
\begin{equation}
    \part{\L}{\mu_k} = \rbar{\part{\L}{\mu_k}}_{\hat{\mu}'_k} + \rbar{\part{\L}{\mu_k}}_{J_k}
\end{equation}

The above gradients were incorporated into a PyTorch automatic differentiation framework, based largely on the reference implementation provided by \cite{kerbl20233d}.
\end{appendices}

\end{document}